\documentclass{article} 
\usepackage[preprint]{colm2026_conference}
\usepackage{microtype}
\usepackage{hyperref}
\usepackage{url}
\usepackage{booktabs}

\usepackage{rotating}
\usepackage{threeparttable}
\usepackage{amsmath,amsfonts,bm}

\def \real    { \mathbb{R} }
\def \eg      {e.g.}

\def\eqref#1{equation~\ref{#1}}

\def\1{\bm{1}}

\DeclareMathAlphabet{\mathsfit}{\encodingdefault}{\sfdefault}{m}{sl}
\SetMathAlphabet{\mathsfit}{bold}{\encodingdefault}{\sfdefault}{bx}{n}

\usepackage{graphicx}
\usepackage{subcaption}

\usepackage{lineno}

\definecolor{darkblue}{rgb}{0, 0, 0.5}
\hypersetup{colorlinks=true, citecolor=darkblue, linkcolor=darkblue, urlcolor=darkblue}

\title{TLoRA+: A Low-Rank Parameter-Efficient Fine-Tuning Method for Large Language Models}

\author{
  Yarui Cao \\
  Department of Computer Science \\
  Clemson University \\
  Clemson, SC 29634, USA \\
  \texttt{yaruic@clemson.edu} \\
  \And
  Kai Liu \\
  Department of Computer Science \\
  Clemson University \\
  Clemson, SC 29634, USA \\
  \texttt{kail@clemson.edu}
}

\begin{document}

\ifcolmsubmission
\linenumbers
\fi

\maketitle

\begin{abstract}
Fine-tuning large language models (LLMs) aims to adapt pre-trained models to specific tasks using relatively small and domain-specific datasets. Among Parameter-Efficient Fine-Tuning (PEFT) methods, Low-Rank Adaptation (LoRA) stands out by matching the performance of full fine-tuning while avoiding additional inference latency. In this paper, we propose a novel PEFT method that incorporates the TLoRA+ optimizer into the weight matrices of pre-trained models. The proposed approach not only preserves the efficiency of low-rank adaptation but also further enhances performance without significantly increasing computational cost. We conduct experiments on the GLUE benchmark across diverse model architectures. Numerical experiments consistently demonstrate the effectiveness and robustness of our proposed method.
\end{abstract}

\section{Introduction}
Fine-tuning large language models (LLMs)~\citep{Hoffmann2022TrainingCL} aims to adapt a pre-trained model on a smaller, domain-specific dataset to perform specific tasks or improve its performance. This process refines the model's weights to enhance task performance, inject desirable behaviors, and eliminate undesirable ones. However, fine-tuning very large models requires significant computational resources. For example, fine-tuning a 70B-parameter LLaMA3 model requires around 500GB of GPU memory. To address these challenges, various methods have been proposed to reduce the number of trainable parameters and memory usage. Parameter-Efficient Fine-Tuning (PEFT)~\citep{Xu2023ParameterEfficientFM} has become the most popular method, making large models efficiently adapt to various downstream tasks without fine-tuning all parameters. By training only a small subset of parameters, we can significantly reduce computational and storage costs while achieving performance comparable to that of a fully fine-tuned model. This makes it more accessible for researchers to fine-tune LLMs with limited hardware resources. Among these PEFT methods, Low-Rank Adaptation (LoRA)~\citep{Hu2021LoRALA} stands out by maintaining the performance of full fine-tuning without introducing additional inference latency, making it a highly efficient technique for adapting large models.

\begin{figure*}
    \centering
    \includegraphics[width=0.8\linewidth]{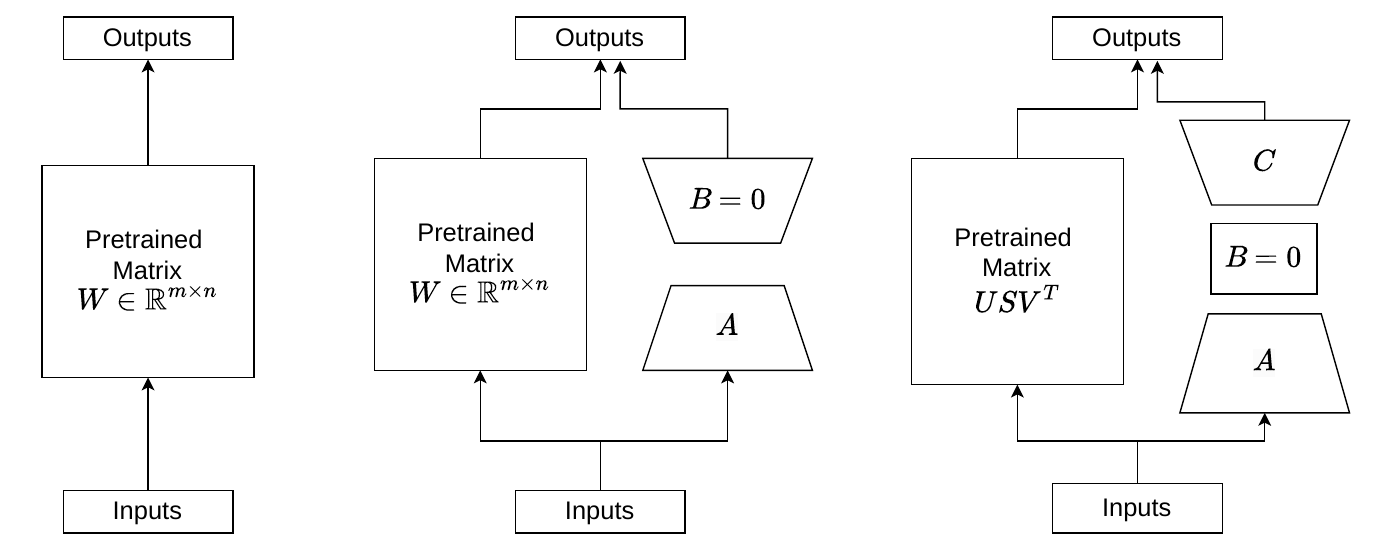}
    \caption{Structural comparison of adaptation strategies (left to right): full fine-tuning, LoRA and TLoRA.}
    \label{fig:lora structure}
\end{figure*}

The basic idea of LoRA is to design low-rank matrices that are then added to the original matrix, enhancing its structure without significantly increasing time consumption. As Figure~\ref{fig:lora structure} shows, for a pre-trained weight matrix $W_0 \in \mathbb{R}^{m\times n}$, LoRA substitutes the updates with a low-rank decomposition $\Delta W = BA$, where $B \in \mathbb{R}^{m\times r}$ and $A \in \mathbb{R}^{r\times n}$, and the rank $r \ll \min (m, n)$. For $h = W_0 x$, the modified forward pass yields:
\begin{equation}
\label{eq:h}
    h = (W_0 + \Delta W)x = W_0x + BAx.
\end{equation}

A random Gaussian initialization is applied to $A$ and zero to $B$, making $BA = 0$ at the beginning of the training. As a result, injecting the adapter doesn't initially affect the model's output. With this design, the need to compute gradients or maintain the optimizer states of the original matrix $W$ can be avoided, allowing us to focus on optimizing low-rank matrices $A$ and $B$. Thus, the goal of reducing memory usage can be achieved. Moreover, LoRA can match or even surpass the performance of full fine-tuning, indicating that fine-tuning only a subset of parameters is sufficient for downstream tasks.


\section{Literature Reviews}
With the explosion of information, LLMs with billions of parameters have shown remarkable performance in specific downstream tasks~\citep{Gadre2024LanguageMS}. PEFT has become a popular method that reduces the number of trainable parameters and memory requirements while maintaining a performance comparable to full fine-tuning. PEFT encompasses several strategies, including partial fine-tuning, soft prompt fine-tuning, non-linear adapter fine-tuning, and low-rank adapter-based fine-tuning.

LoRA injects trainable adapters into linear layers, enabling efficient fine-tuning by re-parameterizing these adaptations into the standard model structure. This method has been widely adopted to maintain the original model architecture while improving fine-tuning efficiency. Based on LoRA, AdaLoRA~\citep{Zhang2023AdaLoRAAB} dynamically allocates the parameter budget among weight matrices according to their importance scores, effectively pruning the singular values of the less important updates. Delta-LoRA~\citep{Zi2023DeltaLoRAFH} addresses LoRA's representational capacity for downstream tasks by leveraging the delta of the product of two low-rank matrices. Incorporating sparsity constraints, SoRA~\citep{Ding2023SparseLA} combines LoRA with sparse updates, enabling dynamic adjustments to the intrinsic rank during the adaptation process to examine the impact of the number of non-zero parameters on the model's memorization and generalization. LoRA+~\citep{Hayou2024LoRAEL} corrects the suboptimality of LoRA by assigning different learning rates to the LoRA adapter matrices A and B. This adjustment improves performance and fine-tuning speed while maintaining a similar computational cost to LoRA. DoRA~\citep{Liu2024DoRAWL} enhances LoRA's learning capacity and training stability by decomposing the pre-trained weights into magnitude and direction components, avoiding additional inference overhead. To address the slow convergence problem, PiSSA~\citep{Meng2024PiSSAPS} introduces principal singular values and singular vectors adaptation to accelerate training. LoRA-GA~\citep{Wang2024LoRAGALA} proposes a novel initialization method that aligns the gradients of the low-rank matrix product with those of full fine-tuning at the first step. This technique achieves a convergence rate comparable to full fine-tuning while delivering similar or superior performance. HydraLoRA~\citep{Tian2024HydraLoRAAA} introduces an asymmetric structure into LoRA to eliminate the reliance on domain knowledge during training and inference. Finally, NoRA~\citep{Lin2024NoRANL} utilizes a dual-layer nested structure with SVD, extending the capacity of LoRA while reducing the number of tunable parameters. 

Beyond classical two-matrix formulations, matrix factorization research provides a broader foundation for exploring multi-factor decompositions. Traditional methods such as SVD, QR decomposition, and various tri-matrix factorizations that decompose a matrix into multiple structured components can enhance interpretability and flexibility. In particular, tri-factor models separate scaling, basis, and transformation components, offering improved adaptability in representation learning. Motivated by these advantages, TLoRA~\citep{Islam2025TLoRATL} extends the standard LoRA framework from a two-matrix to a three-matrix decomposition as shown in Figure~\ref{fig:lora structure}, aiming to enhance expressive capacity while preserving the efficiency of low-rank adaptation. By introducing an additional trainable transformation matrix and maintaining the other two factors fixed, TLoRA achieves highly efficient parameter adaptation with only minimal computational overhead. This tri-factorization low-rank adaptation approach has also been adopted in personalized model parameter aggregation, as demonstrated by CE-LoRA~\citep{li2025communication}, which significantly reduces communication cost while maintaining comparable empirical performance.


\section{Methodology}

\subsection{Tri-Matrices Adapter}
\label{sec: tri-matrix}
In this section, we explore TLoRA and its variants. Compared to the standard LoRA, TLoRA employs a tri-matrix decomposition to compute the weight update $\Delta W \in \real^{m \times n}$. Specifically, the update is parameterized by three low-rank matrices: $C \in \mathbb{R}^{m \times r_1}$, $B \in \mathbb{R}^{r_1 \times r_2}$, and $A \in \mathbb{R}^{r_2 \times n}$, where $r_1, r_2 \ll \min(m, n)$. The resulting low-rank update is given by $\Delta W = CBA$. Accordingly, the modified forward pass can be expressed as:
\begin{equation}
h = (W_0 + \Delta W)x = W_0x + CBAx.
\end{equation}

Unlike LoRA and the variants discussed earlier, only the matrix $B$ is trainable, while $A$ and $C$ are randomly initialized and fixed during the adaptation. 

This design preserves the computational efficiency of low-rank updates while avoiding the additional parameter growth introduced by learning multiple factors. However, fixing part of the decomposition may also reduce the expressive capacity of the adapter, potentially limiting the amount of task-relevant information it can capture. To better understand this trade-off, we investigate several configurations of the tri-factor parameterization:
\begin{enumerate}
\item Only $B$ is trainable, which is the original setting of TLoRA. This is the most parameter-efficient setting. The adaptation is governed only by the learned scaling matrix $B$, while $A$ and $C$ provide fixed subspace projections.
\item $A(\text{or }C)$ and $B$ are trainable. Allowing a single projection matrix to be updated increases flexibility and brings the adapter closer to the standard LoRA.
\item All three matrices are trainable. This setting provides the highest expressive power and adapts the entire low-rank subspace, but also introduces the largest number of parameters and computational cost among these three setups.
\end{enumerate}

\begin{figure}
    \centering
    \includegraphics[width=0.95\linewidth]{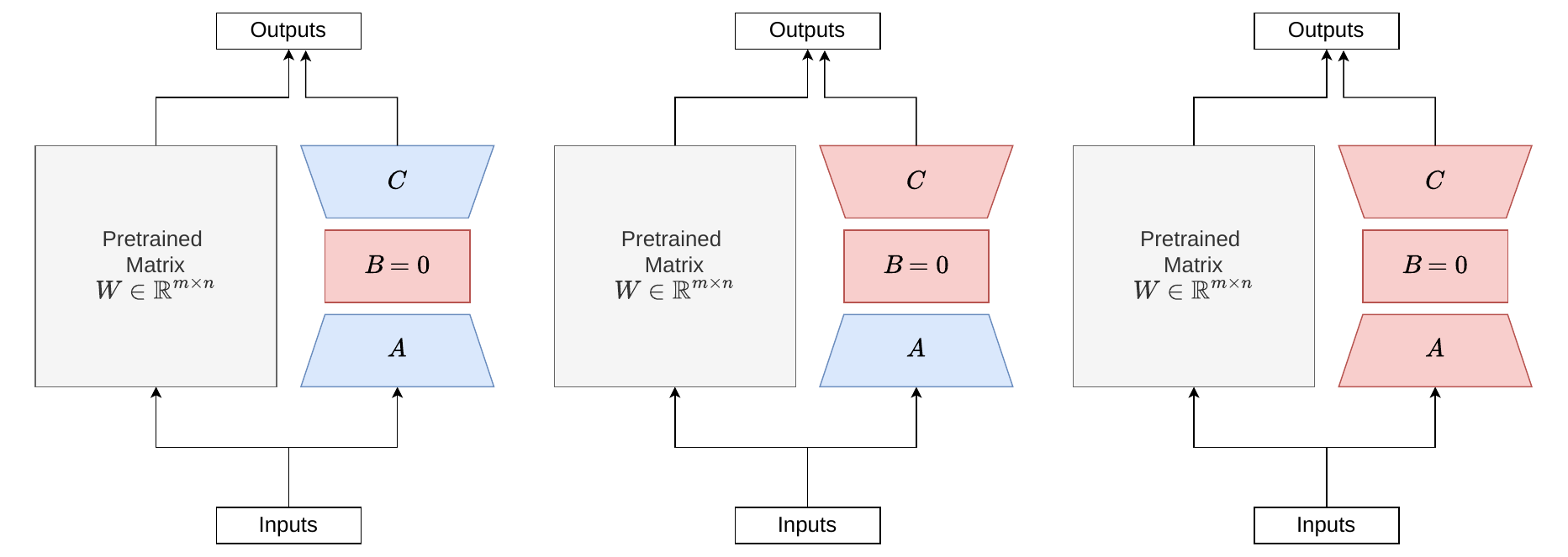}
    \caption{Three configurations of the tri-matrices adapter. Red indicates trainable matrices, while blue denotes frozen (non-trainable) matrices. }
    \label{fig:Tlora}
\end{figure}

The initialization setup is visualized in Figure~\ref{fig:Tlora}. These configurations are designed to characterize the trade-off between parameter efficiency and representational capacity in tri-factorization low-rank adaptation.

\subsection{TLoRA+ Optimizer}
Drawing inspiration from the learning rate adjustments for matrices $A$ and $B$ proposed by~\citet{Hayou2024LoRAEL}, we extend this formulation as TLoRA+ to accommodate our three-matrix framework.

Following the findings from LoRA+ that optimal learning rate adjustments for trainable matrices are independent of the pre-trained weights, we assume $W_0 = 0$ with loss of generality. We denote the forward pass as $Y = CBAX$, where $X \in \real^{n \times b}$ represents the input. Based on the LeCun initialization~\citep{Hartmanis2002NeuralNT}, if the weight matrix $W \in \mathbb{R}^{m \times n}$ is sampled i.i.d.\ from a distribution with mean $0$ and variance $\frac{1}{n}$, the components of the product $WX$ will have roughly the same expected magnitude as the components of $X$. Applying this principle, if the input $X$ is of order $\mathcal{O}(1)$, then matrices $A, B$ and $C$ must be initialized with variances $\frac{1}{n}, \frac{1}{r_2}$ and $\frac{1}{r_1}$, respectively, to ensure that the components of all $AX, BAX$ and $CBAX$ remain of order $\mathcal{O}(1)$.

For a given loss function $L$, the first-order approximation of the change in loss is as follow:
\begin{equation}
\label{eq:loss}
    L(A + \Delta A, B + \Delta B, C + \Delta C) - L(A, B, C)
    \approx \langle \frac{\partial L}{\partial A}, \Delta A \rangle 
    + \langle \frac{\partial L}{\partial B}, \Delta B \rangle
    + \langle \frac{\partial L}{\partial C}, \Delta C \rangle,
\end{equation}
where $\frac{\partial L}{\partial A}$, $\frac{\partial L}{\partial B}$ and $\frac{\partial L}{\partial C}$ denote the gradient of the loss with respect to matrices $A, B$ and $C$, respectively. And $\langle \cdot, \cdot \rangle$ denotes the Frobenius inner product.

Consider the Adam optimizer in the limiting case where the hyperparameters $\beta_1$ and $\beta_2$ are both set to zero. Under this setting, Adam degenerates into SignSGD~\footnote{AdamW, in contrast, reduces to SignSGD with decoupled weight decay.}, and the parameter updates for the matrices are
\begin{equation}
\label{eq:gradient}
    \Delta A = -\eta_A \text{sign}(\frac{\partial L}{\partial A}), \quad
    \Delta B = -\eta_B \text{sign}(\frac{\partial L}{\partial B}), \quad
    \Delta C = -\eta_C \text{sign}(\frac{\partial L}{\partial C}),
\end{equation}
where $\eta_A, \eta_B$ and $\eta_C$ are the learning rates associated with matrices $A, B$ and $C$, respectively.

Substituting the update rules from Eq.~(\ref{eq:gradient}) into Eq.~(\ref{eq:loss}), then
\begin{equation}
    L(A + \Delta A, B + \Delta B, C + \Delta C) - L(A, B, C)
    \approx \underbrace{-\eta_A || \frac{\partial L}{\partial A} ||_1}_{\Delta L_A}
    \underbrace{-\eta_B || \frac{\partial L}{\partial B} ||_1}_{\Delta L_B}
    \underbrace{-\eta_C || \frac{\partial L}{\partial C} ||_1}_{\Delta L_C},
\end{equation}
where $|| \cdot ||_1$ denotes $\ell_1$-norm. Under the assumption that each matrix should contribute equally to the reduction of the loss during each update, the terms $\Delta L_A, \Delta L_B$ and $\Delta L_C$ on the right-hand side should be of the same order of magnitude.

Taking the derivative of $A, B$ and $C$ with respect of $Y$, we have:
\begin{align}
    \frac{\partial L}{\partial A} 
    &= \frac{\partial L}{\partial Y} \frac{\partial Y}{\partial A} 
    = B^T C^T \frac{\partial L}{\partial Y} X^T, \\ \nonumber
    \frac{\partial L}{\partial B}
    &= \frac{\partial L}{\partial Y} \frac{\partial Y}{\partial B} 
    = C^T \frac{\partial L}{\partial Y} X^T A^T, \\ \nonumber
    \frac{\partial L}{\partial C}
    &= \frac{\partial L}{\partial Y} \frac{\partial Y}{\partial C} 
    = \frac{\partial L}{\partial Y} X^T A^T B^T. 
\end{align}

The term $\Delta L_A$ is proportional to the $\ell_1$-norm of the gradient, $|| \frac{\partial L}{\partial A} ||_1$, which is the sum of the absolute values of its $nr_2$ components. Assuming these components are of comparable magnitude, $\Delta L_A$ scales approximately linearly with $nr_2$. Furthermore, because $\frac{\partial L}{\partial A}$ is linear with respect to $B^T C^T$, we can generally assume the magnitude of each component of $\frac{\partial L}{\partial A}$ is proportional to the overall magnitude of $B^T C^T$. Consequently, $\Delta L_A$ is proportional to both $nr_2$ and the magnitude of $B^T C^T$. By similar logic, $\Delta L_B$ is approximately proportional to both $r_1 r_2$ and the magnitude of $C^T A^T$, while $\Delta L_C$ is approximately proportional to both $mr_1$ and the magnitude of $A^T B^T$. Then, we have
\begin{align*}
    \Delta L_A \approx &\Delta L_B \approx \Delta L_C \\
    \eta_A nr_2 \sqrt{\frac{1}{r_1 r_2}} \approx \eta_B r_1r_2 &\sqrt{\frac{1}{n r_1}} \approx \eta_C mr_1 \sqrt{\frac{1}{nr_2}} \\
    \Rightarrow \eta_A : \eta_B \approx \frac{r_1\sqrt{r_2}}{n\sqrt{n}}, \quad 
     &\eta_B : \eta_C \approx \frac{m\sqrt{r_1}}{r_2\sqrt{r_2}}.
\end{align*}

To simplify the analysis, we assume $r_1 = r_2 = r$ and $r = \mathcal{O}(1)$. Under these assumptions, the relationship yields:
\begin{equation}
\label{eq:ratio_pre}
    \eta_A : \eta_B : \eta_C \approx 1 : n^{3/2} : m^{-1}n^{3/2}.
\end{equation}

In common transformer architectures, weight matrices can be broadly categorized into three types based on their aspect ratios in MLP and attention layers:

\begin{enumerate}
    \item \textbf{Tall matrices} ($m > n$): The MLP up-projection, typically with $m/n \approx 4$.
    \item \textbf{Wide matrices} ($m < n$): The MLP down-projection, typically with $m/n \approx 1/4$.
    \item \textbf{Square matrices} ($m = n$): The attention projections.
\end{enumerate}

These patterns suggest that attention layers are generally square matrices, while MLP layers exhibit relatively stable and structured aspect ratios across models. Based on this regularity, we assume that the contribution of each layer to the overall adaptation effect is proportional to its number of parameters. Thus, Eq.~(\ref{eq:ratio_pre}) can be simplified as:
\begin{equation}
\label{eq:ratio}
    \eta_A : \eta_B : \eta_C \approx 1 : n^{3/2} : n^{1/2}.    
\end{equation}

This result aligns with the strategy proposed in LoRA+, which dictates that the learning rate for matrix $B$ should be set significantly higher than that of matrix $A$.

\section{Experiments and Results}

In this section, we evaluate the performance of our proposed method through a series of comparative experiments. Our evaluation is structured around two primary tasks:
\begin{enumerate}
    \item Comparative Analysis: Benchmarking our approach against the baselines---standard LoRA and TLoRA.
    \item Optimizer Efficiency: Investigating the performance gains provided by the proposed TLoRA+ optimizer in terms of convergence speed and accuracy.
\end{enumerate}

\begin{table}[b]
\centering
\begin{threeparttable}
\resizebox{\linewidth}{!}{%
\begin{tabular}{cccccccccc}
\toprule
ID & Time(s) & Dataset & Warmup & WD & Tr Acc & Tr Loss & Val Acc & Val Loss & Val MCC \\
\midrule
1 & 542 & RTE & 0.15 & 0.05 & 0.9590 & 0.1167 & 0.8375 & 0.6052 & 0.6739 \\
2 & \textbf{540} & RTE & 0.10 & 0.10 & 0.9502 & 0.1291 & 0.8520 & 0.5770 & 0.7132 \\
3 & 542 & RTE & 0.10 & 0.05 & 0.9555 & 0.1302 & 0.8556 & 0.5545 & 0.7102 \\
4 & \textbf{540} & RTE & 0.15 & 0.10 & 0.9549 & 0.1231 & \textbf{0.8592} & 0.5451 & \textbf{0.7174} \\
5 & 395 & MRPC & 0.10 & 0.05 & 0.9447 & 0.1488 & 0.8946 & 0.3642 & 0.7533 \\
6 & 394 & MRPC & 0.15 & 0.10 & 0.9496 & 0.1358 & 0.8971 & 0.3710 & 0.7581 \\
7 & \textbf{392} & MRPC & 0.15 & 0.05 & 0.9445 & 0.1395 & 0.8995 & 0.3990 & 0.7637 \\
8 & \textbf{392} & MRPC & 0.10 & 0.10 & 0.9465 & 0.1387 & \textbf{0.9020} & 0.3703 & \textbf{0.7709} \\
\bottomrule
\end{tabular}%
}
\end{threeparttable}
\caption{Hyperparameter search results for RoBERTa-large-MNLI (learning Rate: $1 \times 10^{-4}$).}
\label{tab:hyperparams_roberta}
\end{table}

\subsection{Datasets}
We fine-tune selected models on the General Language Understanding Evaluation (GLUE) benchmark~\citep{Wang2018GLUEAM}. This benchmark covers a diverse set of natural language understanding (NLU) tasks, including sentiment analysis (SST-2), linguistic acceptability (CoLA), natural language inference (QNLI, RTE), and paraphrase detection (MRPC).

\subsection{Experimental Setting}
The experiments are conducted on a single NVIDIA H100 80GB GPU. For the first task, we adopt the AdamW optimizer with a batch size of 16 and train for 30 epochs, inserting adapters into all linear layers of the base model. The adapter rank is varied over $[8, 16, 32, 64]$ to evaluate its impact. For the second task, we maintain the same batch size and number of epochs to ensure a fair comparison. In this configuration, we sweep the ratio, $n$ in Eq.~(\ref{eq:ratio}), across the values $[1.0, 2.0, 4.0, 5.0, 8.0, 10.0]$. The remaining hyperparameters are held constant, with the base learning rate set to $5 \times 10^{-5}$, weight decay to 0.1, and the warmup ratio to 0.1.
\begin{figure}[htbp]
	\centering
	\includegraphics[width=\textwidth]{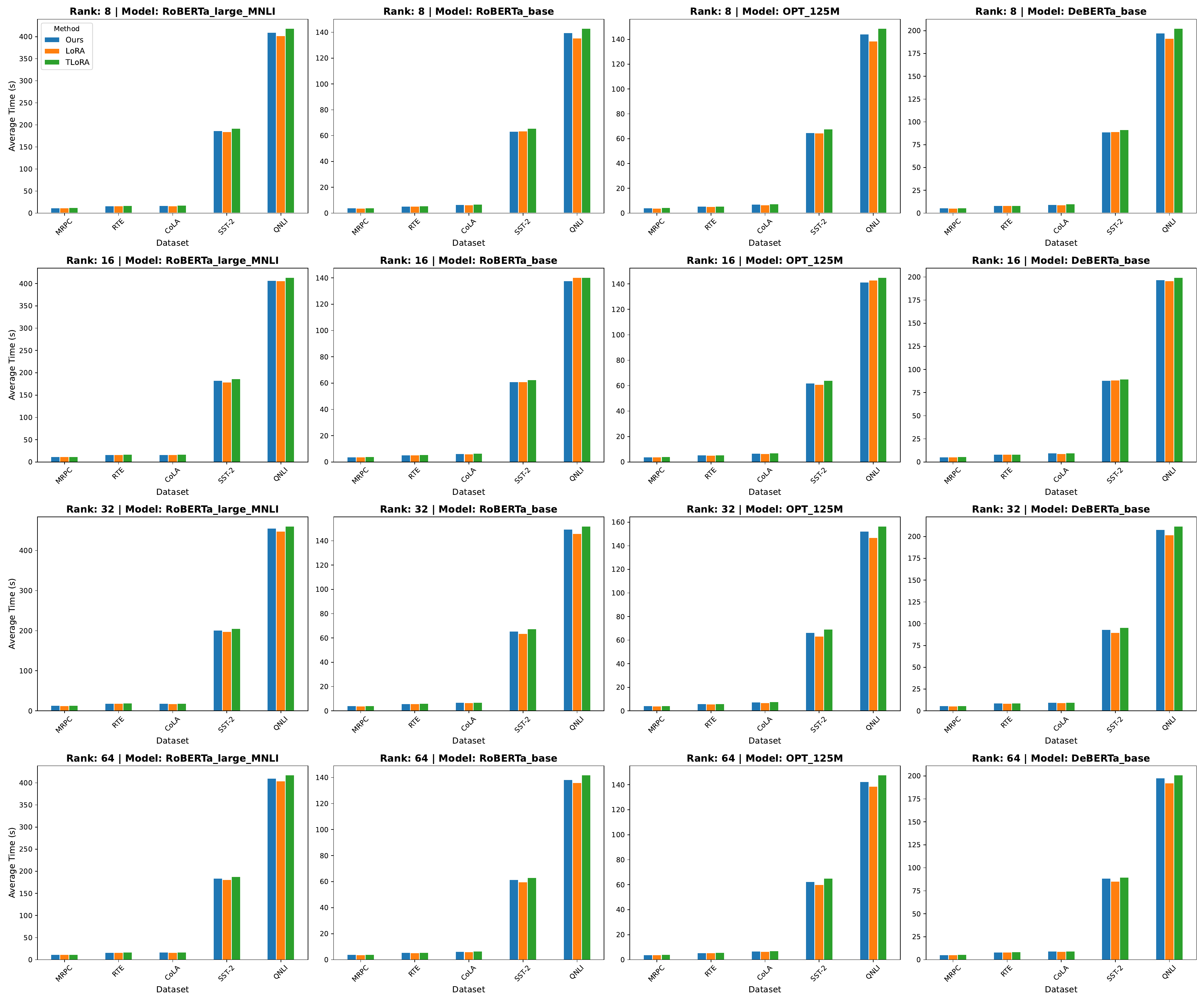}
    \vspace{-2em}
	\caption{Average training time per epoch (in seconds) across five GLUE datasets (MRPC, RTE, CoLA, SST-2 and QNLI) for four base models (RoBERTa-large-MNLI, RoBERTa-base, OPT-125M and DeBERTa-base). The grid organizes models by columns and rank configurations (8, 16, 32 and 64) by rows. Within each subplot, grouped bars compare the computational time of our proposed method, LoRA and TLoRA.}
	\label{fig:time}
\end{figure}

\subsection{Preliminary Experiments}
Before conducting the comparison, we perform preliminary experiments to identify better hyperparameters. Specifically, we fix the rank to a representative value of 16 and explore combinations of learning rates $[1\times10^{-4}, 2\times10^{-4}, 5\times10^{-5}]$, warm-up ratios [0.1, 0.15] and weight decay [0.05, 0.1]. For evaluation, we select the RTE and MRPC datasets from the GLUE benchmark and use RoBERTa-large-MNLI and OPT-125M as the backbone models.

A learning rate of $1\times10^{-4}$ offers an optimal balance between optimization speed and stability, consistently achieving near-convergence in validation loss across warm-up and weight decay configurations. Detailed results are available in Appendix \ref{appendix:lr}.

The hyperparameter combination search results presented in Table~\ref{tab:hyperparams_roberta} indicate that a weight decay (WD) of 0.10 consistently yields strong generalization for the RoBERTa-large-MNLI architecture. 
The combination of a 0.10 warmup ratio and 0.10 weight decay achieves superior computational efficiency compared to other configurations. These findings suggest that, at a learning rate of $1 \times 10^{-4}$, a weight decay of 0.10 is a robust choice for optimizing transformer-based models, while a warmup ratio of 0.10 provides an effective balance between rapid convergence and high predictive performance. Additional results on other architectures are provided in Appendix~\ref{appendix:hyperparameter}.

\begin{figure}[htbp]
	\centering
	\begin{subfigure}{0.48\linewidth}
		\centering
		\includegraphics[width=\linewidth]{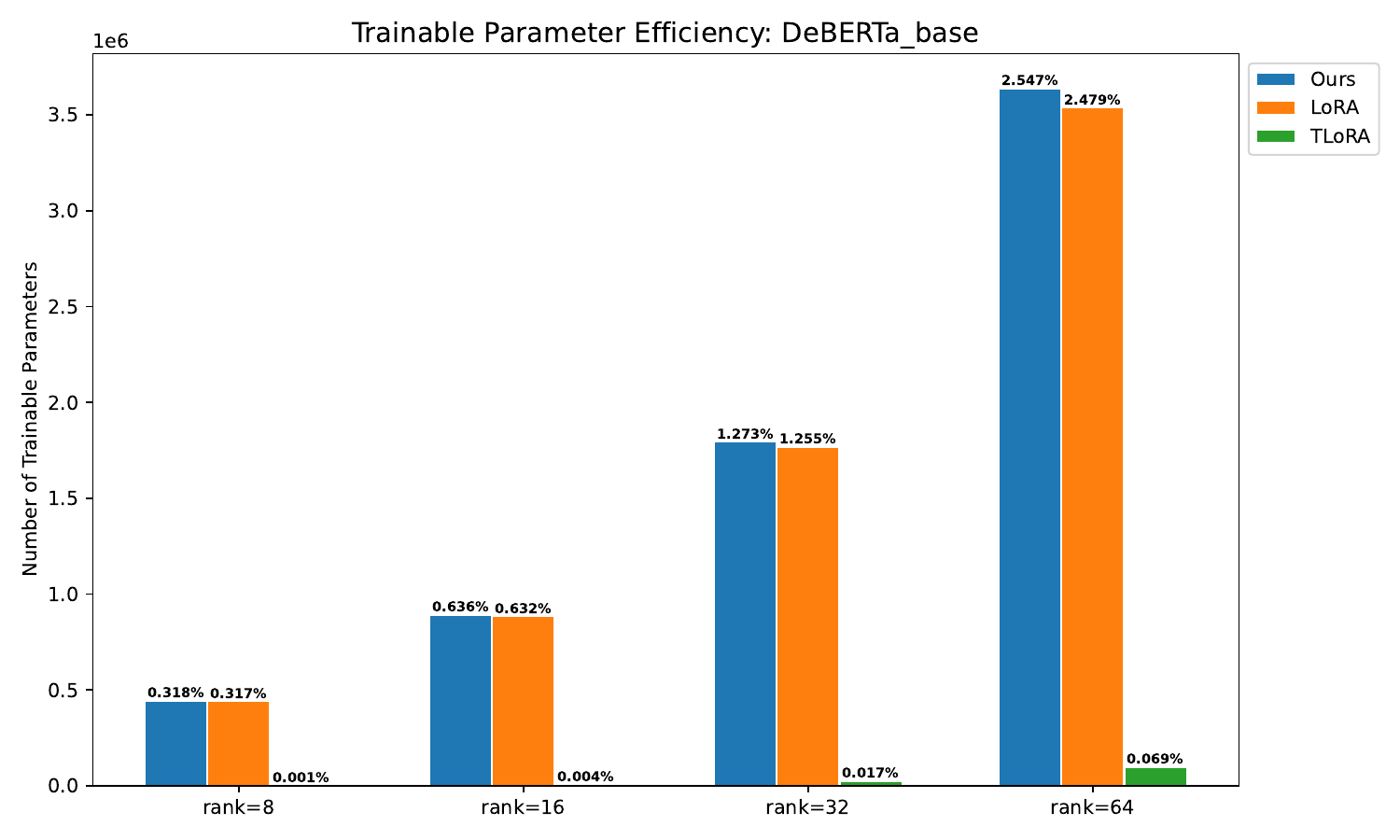}
	\end{subfigure}
	\hfill
	\begin{subfigure}{0.48\linewidth}
		\centering
		\includegraphics[width=\linewidth]{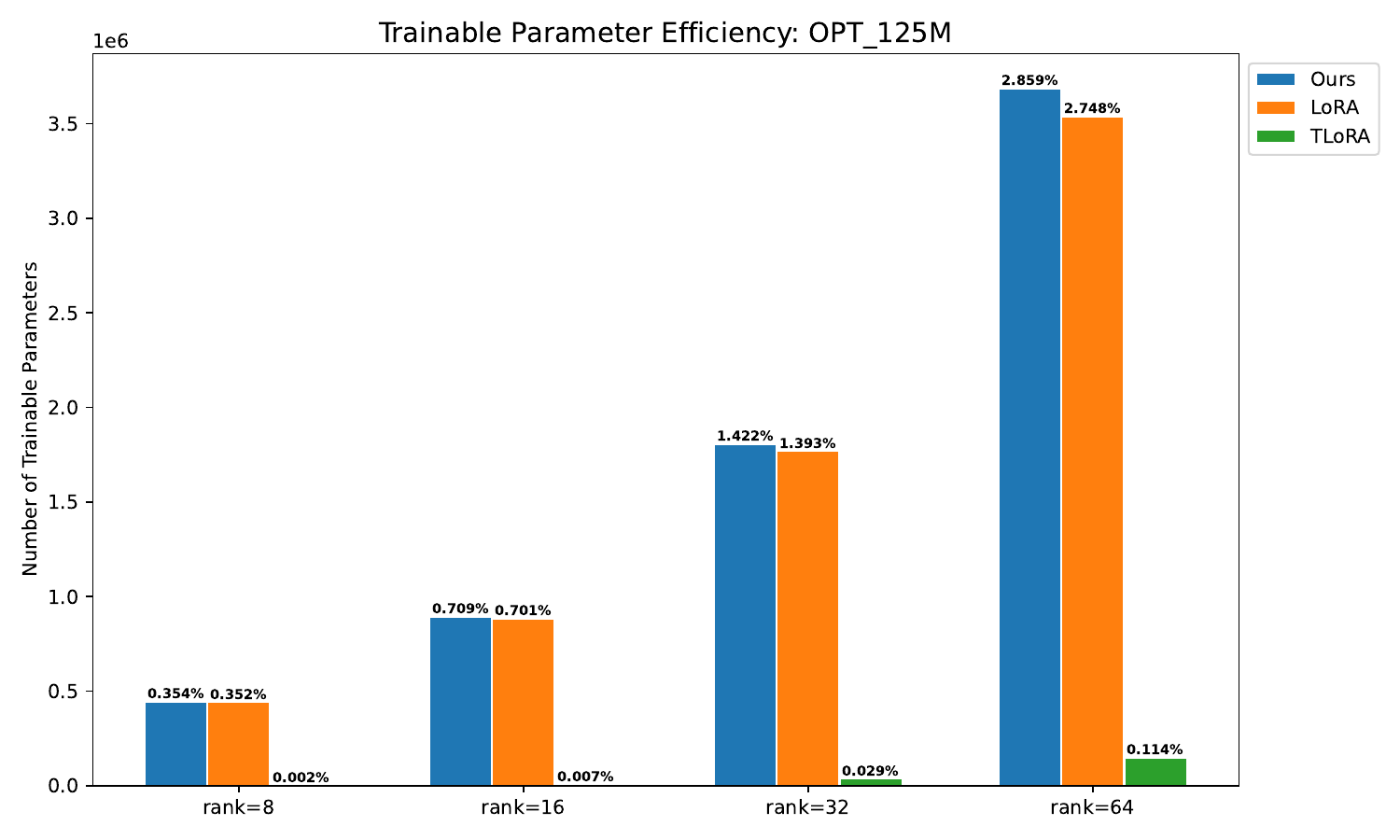}
	\end{subfigure}
	
	\vspace{0.5em}
	
	\begin{subfigure}{0.48\linewidth}
		\centering
		\includegraphics[width=\linewidth]{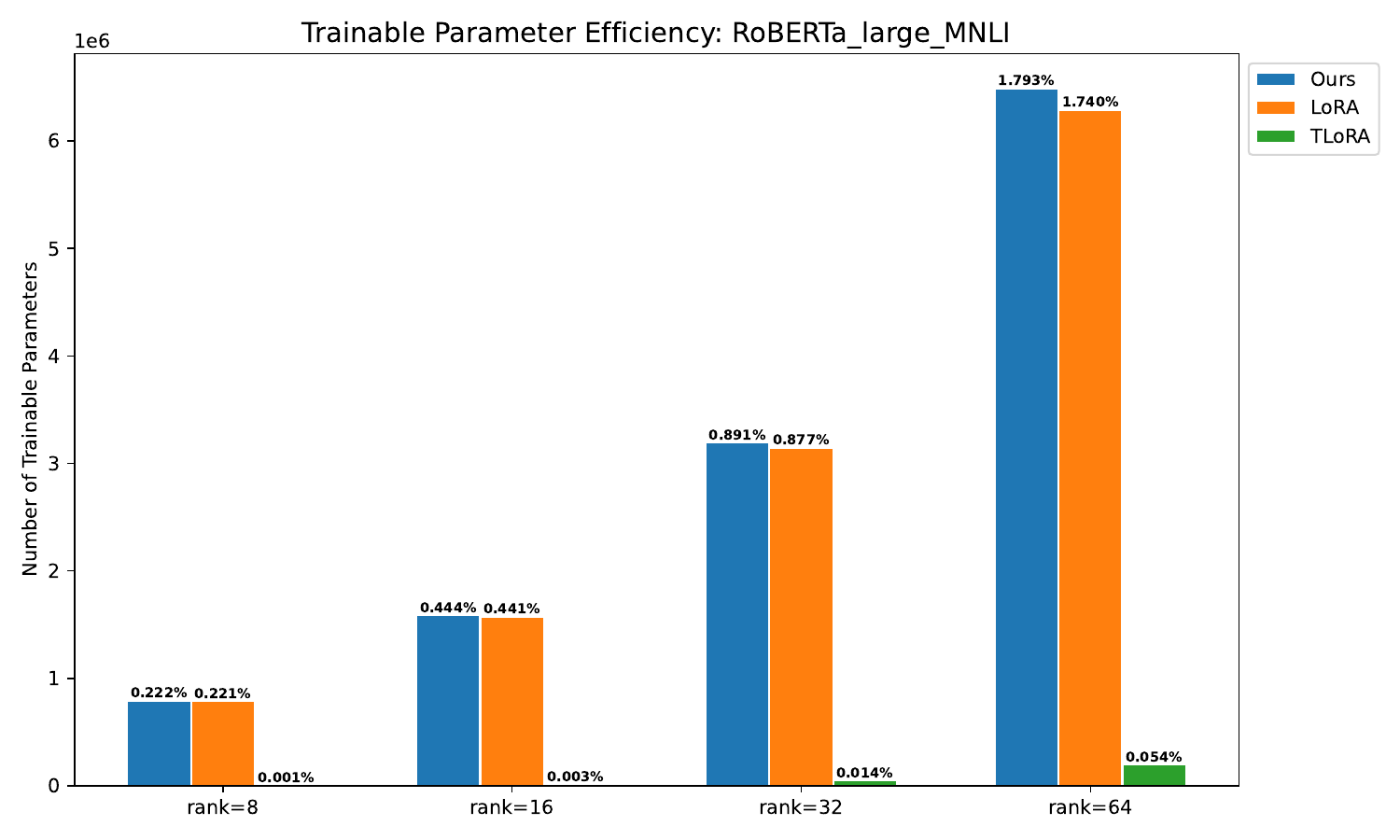}
	\end{subfigure}
	\hfill
	\begin{subfigure}{0.48\linewidth}
		\centering
		\includegraphics[width=\linewidth]{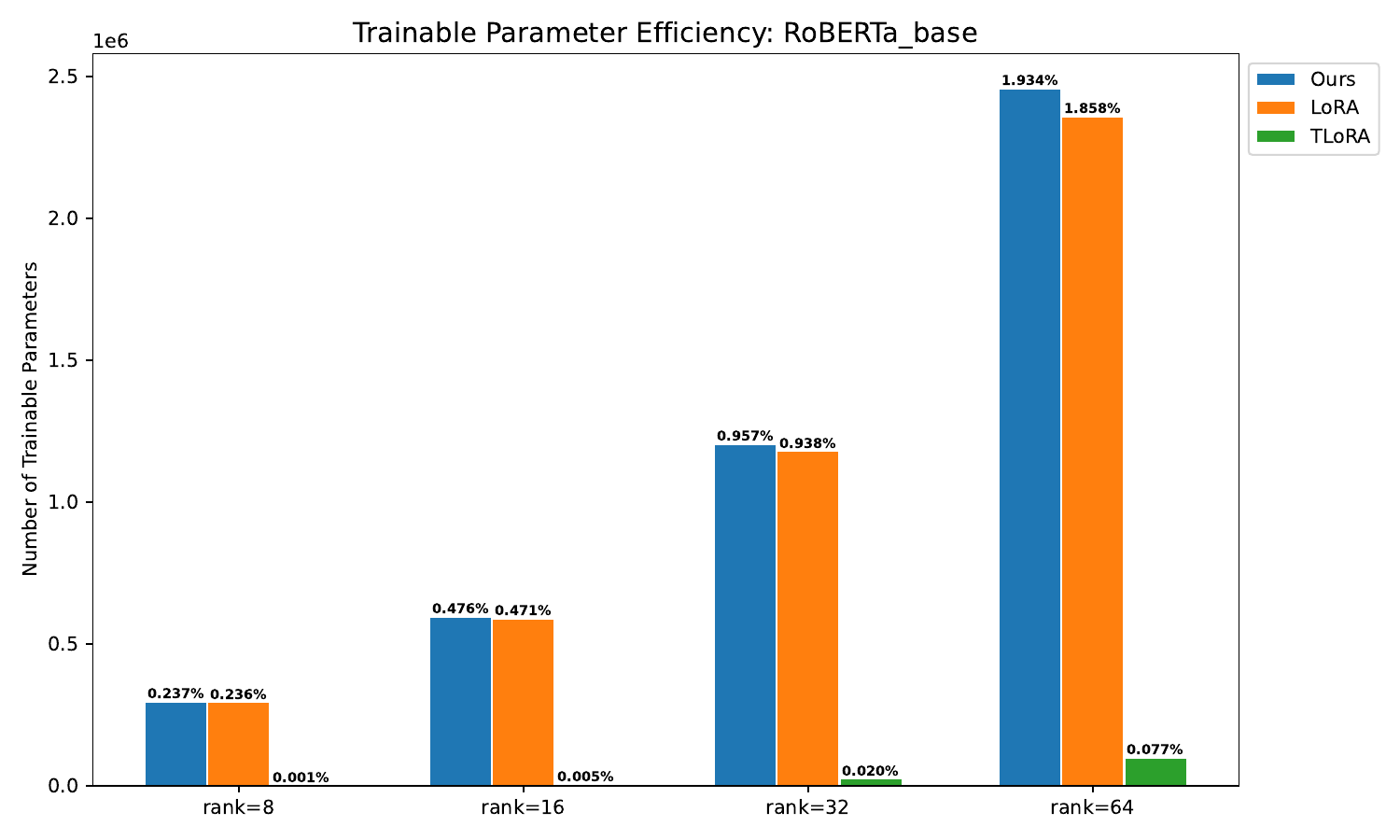}
	\end{subfigure}
	
	\caption{
		Parameter efficiency across four transformer architectures. Each subplot shows the percentage of trainable parameters relative to the total model parameters for ranks 8, 16, 32 and 64, along with the distribution of trainable components for our proposed method, LoRA and TLoRA.
	}
	\label{fig:params}
\end{figure}

To evaluate the trade-off between adaptation flexibility and computational cost, we compare parameter efficiency and memory usage across different architectures (DeBERTa-base, OPT-125M, RoBERTa-large-MNLI and RoBERTa-base) under the three settings introduced in Section~\ref{sec: tri-matrix}. As shown in Figure~\ref{fig:params}, the fraction of trainable parameters increases approximately linearly with rank, while the overall cost remains remarkably small. The TLoRA baseline introduces a negligible number of trainable parameters, whereas our method introduces only a slight increase compared to standard LoRA. 

\subsection{Comparative Analysis}
In this section, we evaluate the performance of standard LoRA, TLoRA and the proposed method across five GLUE benchmark datasets and four transformer backbones. To ensure a fair comparison, all experiments are conducted under a unified hyperparameter setting based on our preliminary experiments, including a warmup ratio of 0.1, a learning rate of $1\times10^{-4}$ and a weight decay of 0.1.
\begin{figure}[t]
    \centering
    \begin{subfigure}[b]{0.9\textwidth} 
        \centering
        \includegraphics[width=\textwidth]{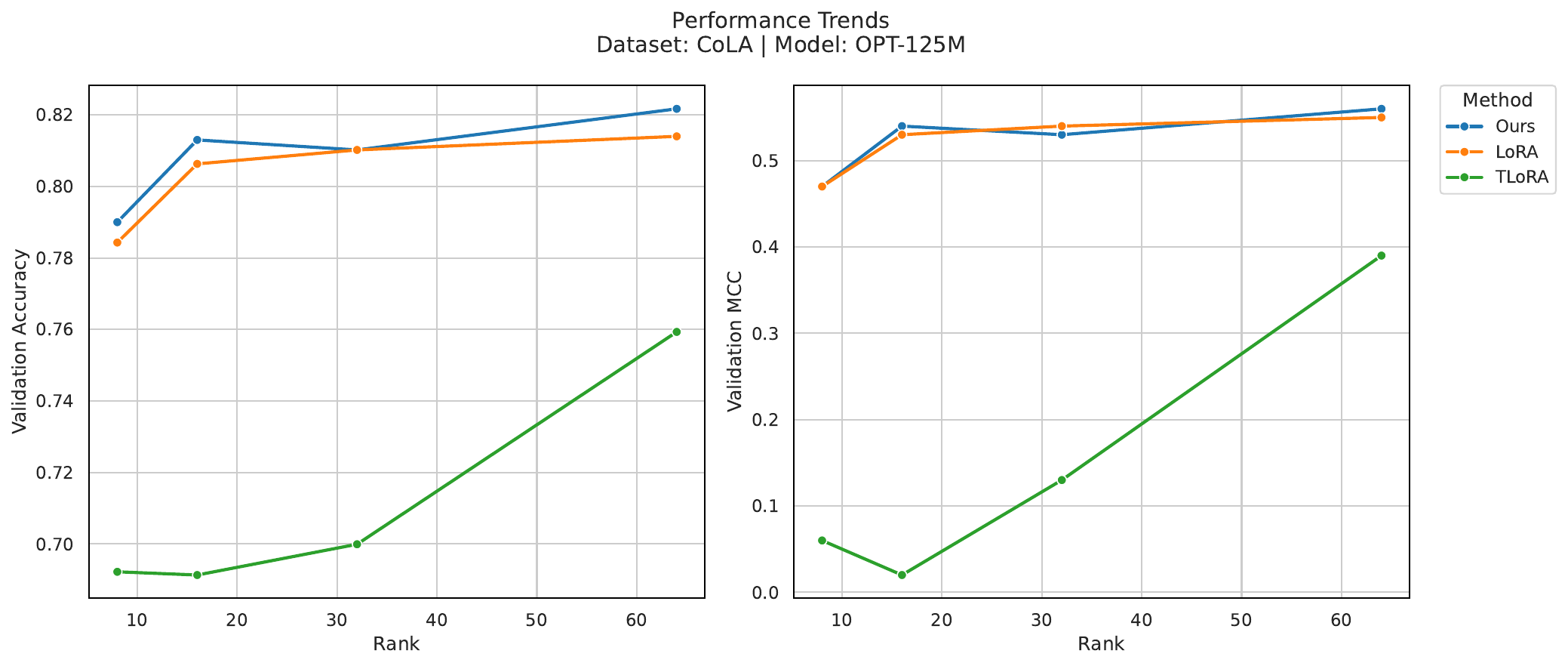}
    \end{subfigure}


    \begin{subfigure}[b]{0.9\textwidth}
        \centering
        \includegraphics[width=\textwidth]{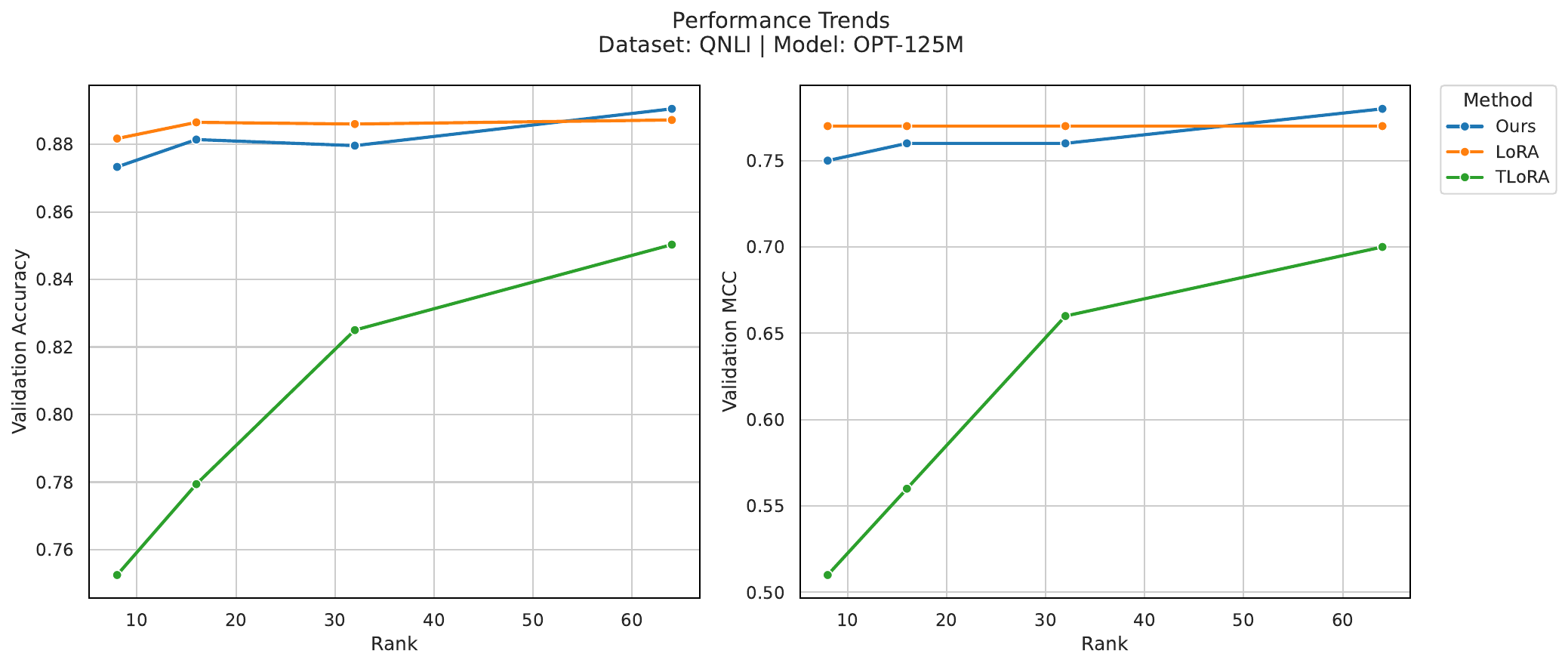}
    \end{subfigure}
    
    \caption{Comparison of trends for CoLA and QNLI datasets under the OPT-125M model.}
    \label{fig:combined_trends}
\end{figure}

Figure~\ref{fig:time} presents the average training time per epoch across the same set of models and rank configurations, evaluated on five datasets. The results indicate that training time is mainly determined by dataset size and model scale, rather than the choice of adapter method or rank. Specifically, larger datasets such as QNLI and SST-2 require longer processing times across all evaluated models, while RoBERTa-large-MNLI is the most computationally demanding in terms of backbone complexity. Additionally, increasing the rank from 8 to 64 leads to only negligible changes in training time. Overall, the runtime performance of LoRA, TLoRA, and our method remains virtually indistinguishable across all configurations.
Combining the observations from Figure~\ref{fig:params}, we note that while the number of trainable parameters in our method increases with rank, the training time remains effectively constant.

Figure~\ref{fig:combined_trends} presents a comparative evaluation of our proposed method against standard LoRA and TLoRA baselines on the CoLA and QNLI datasets using the OPT-125M model. As the rank increases, our method demonstrates strong robustness across both validation accuracy and MCC. While LoRA remains a competitive baseline, our approach achieves its most significant performance gains at higher ranks. In contrast, TLoRA consistently underperforms across all evaluated architectures and rank configurations. Additional results are provided in Appendix~\ref{appendix:rank table}.

Furthermore, the empirical findings highlight varying degrees of architectural sensitivity to the adaptation techniques, with RoBERTa-large-MNLI as the most robust foundation model across all evaluated methods. Due to space limitations, results for additional datasets that exhibit performance trends analogous to those observed on the MRPC dataset are omitted from the main body. Detailed validation accuracies for the QNLI dataset, evaluated across different ranks and models, are provided in Appendix~\ref{appendix:val acc qnli}.

\subsection{Optimizer Efficiency}
This section investigates the trade-off between convergence rate and predictive accuracy by tuning the coefficients among three trainable matrices for rapid and high-precision training. 

\begin{figure}[t]
	\centering
	\begin{subfigure}{0.49\linewidth}
		\centering
		\includegraphics[width=\linewidth]{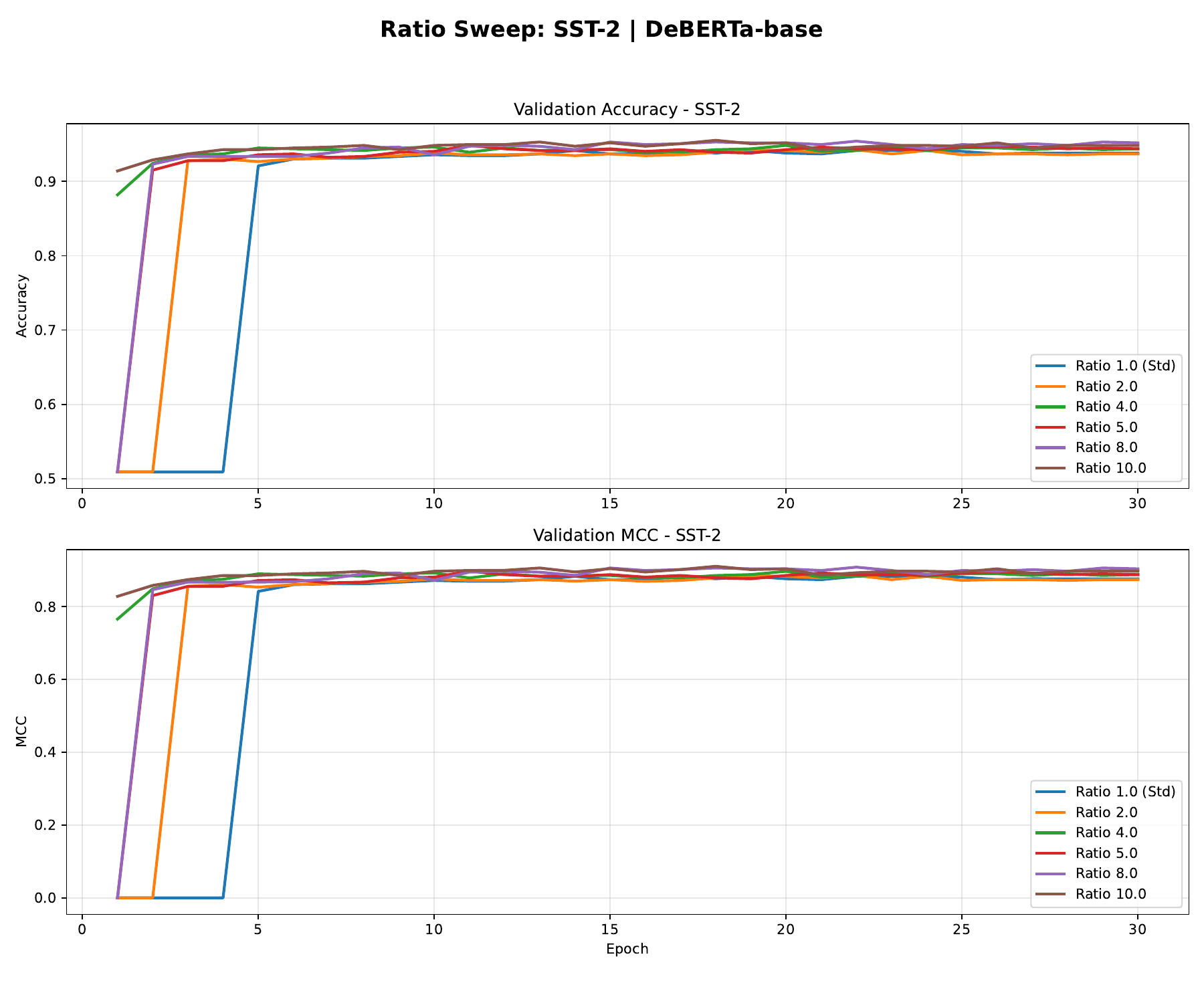}
	\end{subfigure}
	\hfill
	\begin{subfigure}{0.49\linewidth}
		\centering
		\includegraphics[width=\linewidth]{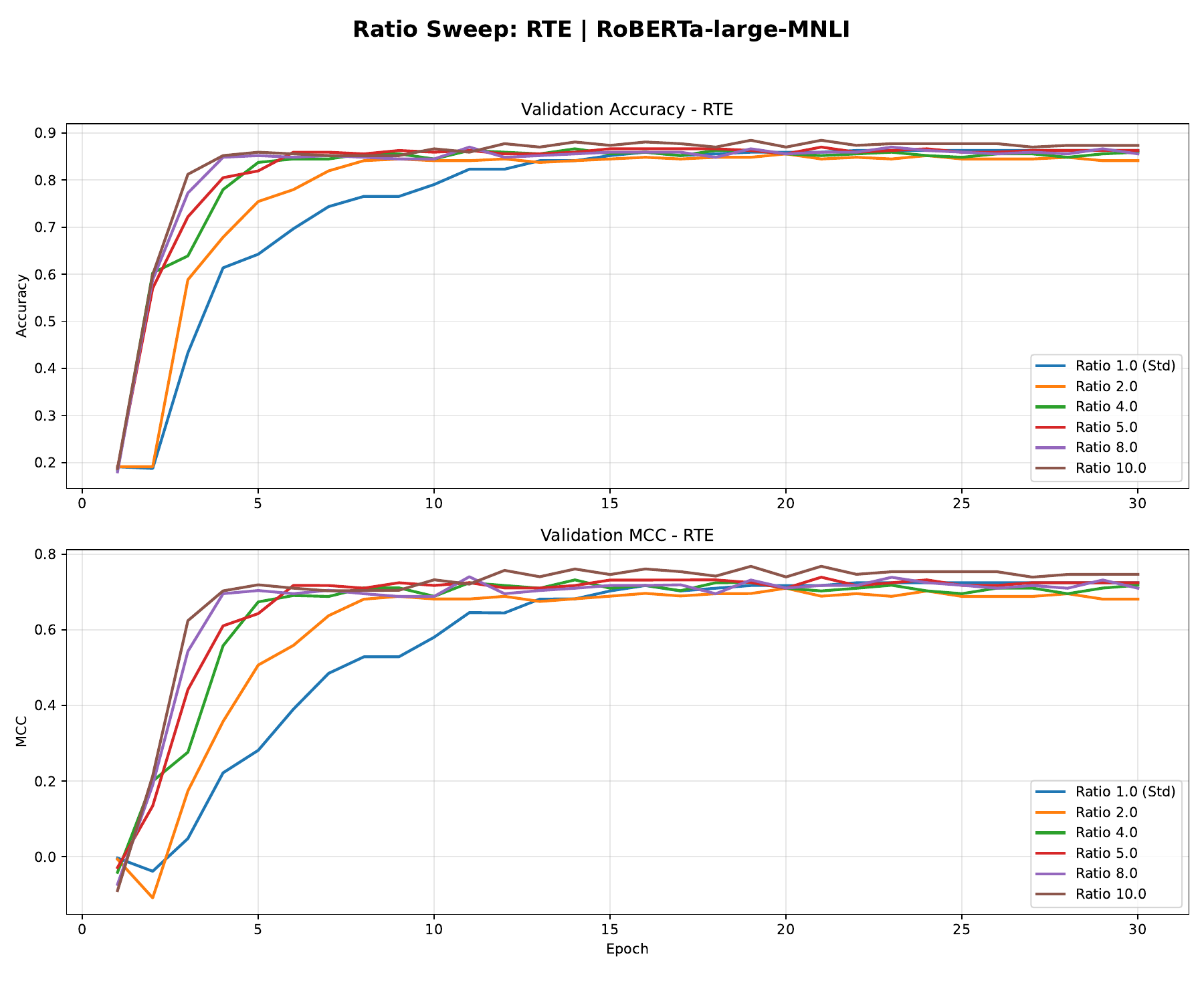}
	\end{subfigure}
	
	\caption{
    		Validation accuracy and MCC on two datasets at rank 8 across different models.
	}
	\label{fig:ratio}
\end{figure}

Partial experimental results across the GLUE benchmark are shown in Figure~\ref{fig:ratio}, which indicates that increasing the ratio beyond the standard 1.0 baseline significantly accelerates convergence and improves performance. This trend is consistent across diverse architectures, including RoBERTa-large-MNLI, RoBERTa-base, OPT-125M, and DeBERTa-base. Measured by validation accuracy and MCC, a higher ratio enables models to reach better performance in fewer epochs overall. The performance gains are most pronounced in the early training epochs, where higher ratios (\eg, 8.0 and 10.0) allow models to bypass the "cold-start" periods often observed at lower ratios in tasks. While the standard ratio 1.0 frequently remains suboptimal, the data suggests a trend of slight fluctuations when exceeding a ratio of 8.0. Therefore, a ratio between 4.0 and 8.0 appears to offer an ideal balance, providing robust generalization and superior convergence efficiency across various NLU tasks. Appendix~\ref{appendix:ratio} provides more experimental results on other datasets and models.

\section{Conclusion}
This paper presents a PEFT method that adopts a structure similar to TLoRA. Experimental results show that it significantly outperforms TLoRA and even achieves better performance than LoRA. In addition, carefully adjusting the learning rates of the three trainable matrices leads to further performance gains. This design enables both efficient fine-tuning and strong parameter efficiency simultaneously. Extensive numerical experiments consistently validate the effectiveness of the proposed approach.

\section{Limitation}
Several questions regarding our method remain unaddressed in this paper. For instance, can incorporating certain constraints further enhance its performance? Can our approach be adapted to convolutional layers to improve performance across various tasks? Additionally, does our method yield similar benefits when combined with quantization? We are actively exploring these questions.

\newpage
\bibliography{colm2026_conference}
\bibliographystyle{colm2026_conference}

\newpage
\appendix

\section{Choice of Learning Rate}
\label{appendix:lr}
\begin{figure}[htbp]
	\centering
	\begin{subfigure}{0.48\linewidth}
		\centering
		\includegraphics[width=\linewidth]{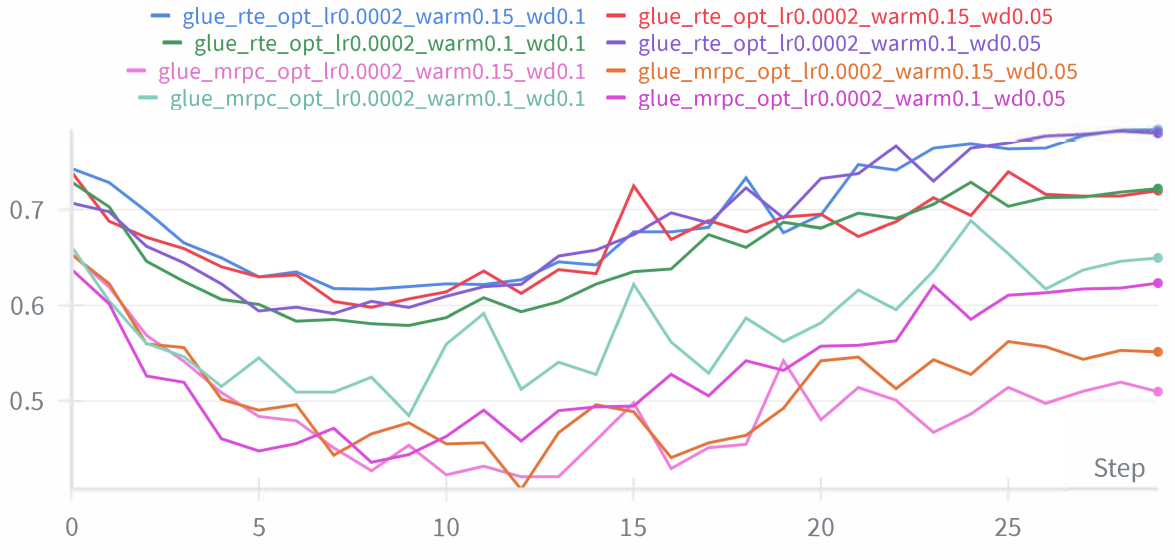}
        \caption{$\text{Learning rate is } 2\times10^{-4}$.}
	\end{subfigure}
	\hfill
	\begin{subfigure}{0.48\linewidth}
		\centering
		\includegraphics[width=\linewidth]{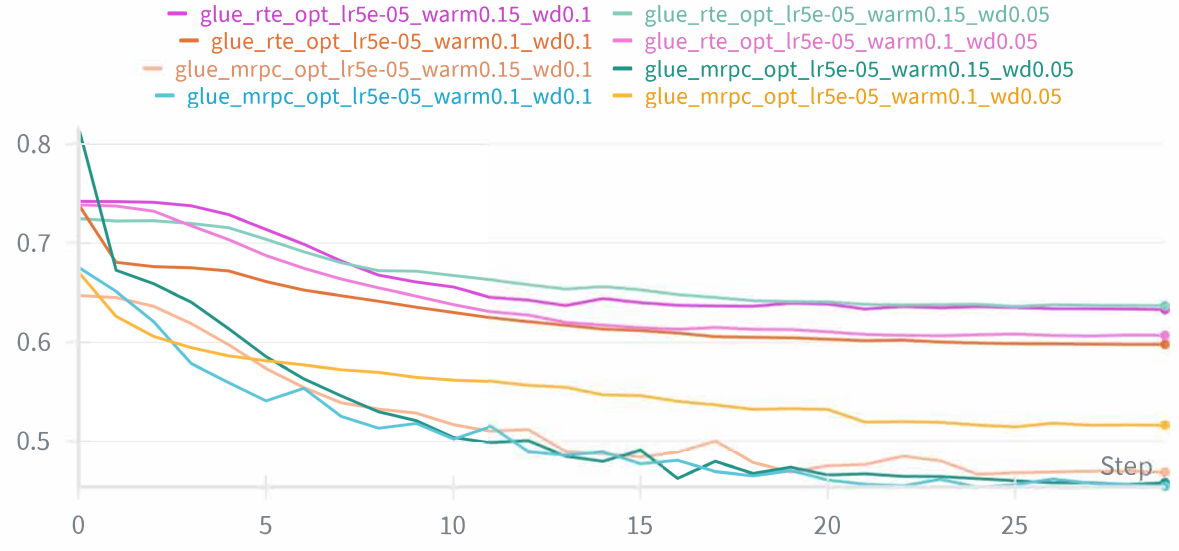}
        \caption{$\text{Learning rate is }5\times10^{-5}$.}
	\end{subfigure}
    
	\vspace{0.5em}
	
	\begin{subfigure}{0.48\linewidth}
		\centering
		\includegraphics[width=\linewidth]{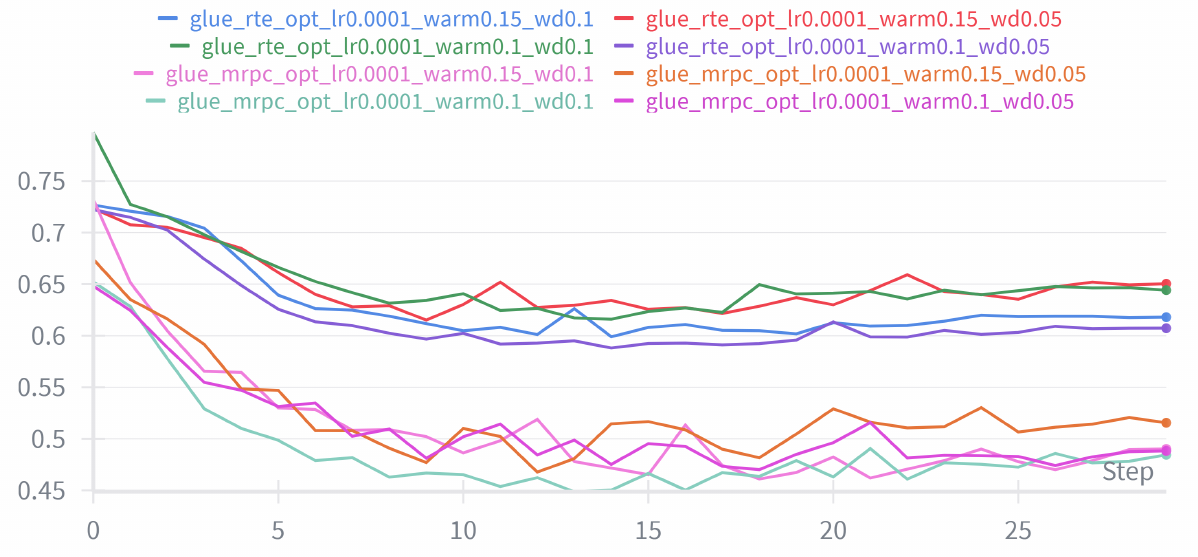}
        \caption{OPT-125M with learning rate $1\times10^{-4}$.}
	\end{subfigure}
	\hfill
	\begin{subfigure}{0.48\linewidth}
		\centering
		\includegraphics[width=\linewidth]{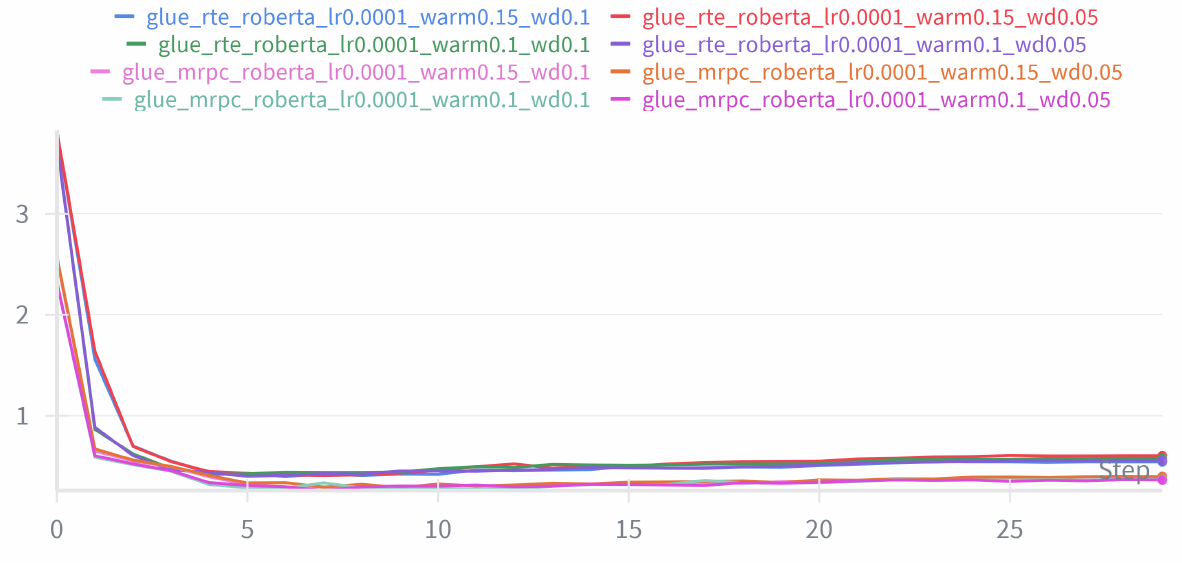}
        \caption{RoBERTa-large-MNLI with learning rate $1\times10^{-4}$.}
	\end{subfigure}
	
	\caption{
        Validation loss under different learning rates and backbone models. A learning rate of $2\times10^{-4}$ leads to overfitting, while $5\times10^{-5}$ results in slow and incomplete convergence. In contrast, $1\times10^{-4}$ achieves stable and near-converged performance across both datasets (RTE and MRPC) and backbone models under all other hyperparameter combinations.
    }
    \label{fig:lr}
\end{figure}

We group the validation loss by learning rate in Figure~\ref{fig:lr} and observe consistent trends across both datasets and backbone models. A learning rate of $2\times10^{-4}$ leads to rapid initial loss reduction but is followed by clear overfitting, indicating overly aggressive updates. In contrast, $5\times10^{-5}$ results in slow optimization and fails to reach convergence within the given training budget. Notably, $1\times10^{-4}$ provides a balanced trade-off between optimization speed and stability, achieving near-converged validation loss across all combinations of warm-up and weight decay. These observations are consistent across both RTE and MRPC datasets, as well as OPT-125M and RoBERTa-large-MNLI backbones.

\section{Hyperparameter Search Results}
\label{appendix:hyperparameter}
\begin{table}[htbp]
\centering
\begin{threeparttable}
\resizebox{\linewidth}{!}{%
\begin{tabular}{cccccccccc}
\toprule
ID & Time(s) & Dataset & Warmup & WD & Tr Acc & Tr Loss & Val Acc & Val Loss & Val MCC \\
\midrule
1 & \textbf{182} & RTE & 0.10 & 0.10 & 0.7786 & 0.5109 & 0.6706 & 0.6441 & 0.3378 \\
2 & 183 & RTE & 0.15 & 0.05 & 0.7729 & 0.4797 & 0.6679 & 0.6504 & 0.3393 \\
3 & \textbf{182} & RTE & 0.10 & 0.05 & 0.7700 & 0.4759 & 0.6751 & 0.6073 & 0.3498 \\
4 & 183 & RTE & 0.15 & 0.10 & 0.7810 & 0.4561 & \textbf{0.6931} & 0.6180 & \textbf{0.3849} \\
5 & 137 & MRPC & 0.15 & 0.05 & 0.8451 & 0.3494 & 0.7892 & 0.5155 & 0.4811 \\
6 & 137 & MRPC & 0.10 & 0.05 & 0.8576 & 0.3320 & 0.7966 & 0.4883 & 0.5006 \\
7 & 137 & MRPC & 0.15 & 0.10 & 0.8453 & 0.3507 & 0.8037 & 0.4902 & 0.5377 \\
8 & \textbf{136} & MRPC & 0.10 & 0.10 & 0.8791 & 0.2859 & \textbf{0.8137} & 0.4845 & \textbf{0.5478} \\
\bottomrule
\end{tabular}%
}
\end{threeparttable}
\caption{Hyperparameter search results for OPT-125M (learning Rate: $1 \times 10^{-4}$).}
\label{tab:hyperparams_opt}
\end{table}

\newpage
\section{Comparison of Three Adaptation Methods at Varying Ranks}
\label{appendix:rank table}
The following tables compare the performance of various models on the MRPC dataset using our method, LoRA and TLoRA across ranks of 8, 16, 32 and 64.

\begin{table}[htbp]
\centering
\resizebox{\textwidth}{!}{
\begin{tabular}{cccccccc}
\toprule
ID & Method & Model & Tr Acc & Tr Loss & Val Acc & Val Loss & Val MCC \\
\midrule
1  & LoRA  & DeBERTa-base       & 0.9918 & 0.0404 & \textbf{0.9044} & 0.3805 & \textbf{0.7747} \\
2  & Ours  & DeBERTa-base       & 0.7045 & 0.4873 & 0.6887 & 0.4780 & 0.1032 \\
3  & TLoRA & DeBERTa-base       & 0.6601 & 0.6762 & 0.6838 & 0.6776 & 0.0000 \\
4  & LoRA  & OPT-125M           & 0.9957 & 0.0125 & \textbf{0.8064} & 1.0821 & \textbf{0.5347} \\
5  & Ours  & OPT-125M           & 0.8062 & 0.4326 & 0.8039 & 0.4529 & 0.5211 \\
6  & TLoRA & OPT-125M           & 0.6785 & 0.6225 & 0.6765 & 0.6100 & 0.0973 \\
7  & LoRA  & RoBERTa-base       & 0.9769 & 0.0763 & \textbf{0.8701} & 0.4543 & \textbf{0.6926} \\
8  & Ours  & RoBERTa-base       & 0.8615 & 0.3214 & 0.8507 & 0.3439 & 0.6324 \\
9  & TLoRA & RoBERTa-base       & 0.6548 & 0.6701 & 0.6838 & 0.6694 & 0.0000 \\
10 & LoRA  & RoBERTa-large-MNLI & 0.9978 & 0.0091 & 0.8997 & 0.5771 & 0.7511 \\
11 & Ours  & RoBERTa-large-MNLI & 0.9246 & 0.1924 & \textbf{0.9044} & 0.3262 & \textbf{0.7753} \\
12 & TLoRA & RoBERTa-large-MNLI & 0.6190 & 0.7626 & 0.5882 & 0.9336 & 0.0837 \\
\bottomrule
\end{tabular}
}
\caption{MRPC rank 8 results.}
\label{tab:mrpc_rank8}

\end{table}

\begin{table}[htbp]
\centering
\resizebox{\textwidth}{!}{
\begin{tabular}{cccccccc}
\toprule
ID & Method & Model & Tr Acc & Tr Loss & Val Acc & Val Loss & Val MCC \\
\midrule
1  & LoRA  & DeBERTa-base       & 0.9818 & 0.0619 & \textbf{0.9020} & 0.3852 & \textbf{0.7693} \\
2  & Ours  & DeBERTa-base       & 0.6832 & 0.4282 & 0.6838 & 0.4160 & 0.1002 \\
3  & TLoRA & DeBERTa-base       & 0.6712 & 0.6379 & 0.6814 & 0.6323 & -0.0337 \\
4  & LoRA  & OPT-125M           & 0.9976 & 0.0093 & \textbf{0.8137} & 1.0633 & \textbf{0.5489} \\
5  & Ours  & OPT-125M           & 0.8684 & 0.3115 & 0.8094 & 0.4804 & 0.5372 \\
6  & TLoRA & OPT-125M           & 0.6768 & 0.6188 & 0.6961 & 0.6144 & 0.1656 \\
7  & LoRA  & RoBERTa-base       & 0.9619 & 0.1027 & \textbf{0.8725} & 0.3828 & \textbf{0.6994} \\
8  & Ours  & RoBERTa-base       & 0.8710 & 0.2942 & 0.8676 & 0.3134 & 0.6966 \\
9  & TLoRA & RoBERTa-base       & 0.6739 & 0.6427 & 0.6838 & 0.6258 & 0.0000 \\
10 & LoRA  & RoBERTa-large-MNLI & 0.9954 & 0.0148 & 0.8995 & 0.5455 & 0.7633 \\
11 & Ours  & RoBERTa-large-MNLI & 0.9483 & 0.1380 & \textbf{0.9022} & 0.3586 & \textbf{0.7665} \\
12 & TLoRA & RoBERTa-large-MNLI & 0.7104 & 0.6065 & 0.7157 & 0.6513 & 0.2455 \\
\bottomrule
\end{tabular}
}
\caption{MRPC rank 16 results.}
\label{tab:mrpc_rank16}

\end{table}

\begin{table}[htbp]
\centering
\resizebox{\textwidth}{!}{
\begin{tabular}{cccccccc}
\toprule
ID & Method & Model & Tr Acc & Tr Loss & Val Acc & Val Loss & Val MCC \\
\midrule
1  & LoRA  & DeBERTa-base       & 0.9628 & 0.1237 & \textbf{0.8922} & 0.3426 & \textbf{0.7465} \\
2  & Ours  & DeBERTa-base       & 0.8851 & 0.2779 & 0.8750 & 0.3072 & 0.7156 \\
3  & TLoRA & DeBERTa-base       & 0.6783 & 0.6099 & 0.6863 & 0.5850 & 0.0729 \\
4  & LoRA  & OPT-125M           & 0.9978 & 0.0083 & 0.8064 & 0.9143 & 0.5300 \\
5  & Ours  & OPT-125M           & 0.8959 & 0.2559 & \textbf{0.8113} & 0.5037 & \textbf{0.5405} \\
6  & TLoRA & OPT-125M           & 0.7205 & 0.5492 & 0.7083 & 0.5371 & 0.2281 \\
7  & LoRA  & RoBERTa-base       & 0.9607 & 0.1179 & 0.8529 & 0.4002 & 0.6511 \\
8  & Ours  & RoBERTa-base       & 0.8919 & 0.2534 & \textbf{0.8824} & 0.3388 & \textbf{0.7249} \\
9  & TLoRA & RoBERTa-base       & 0.6746 & 0.6153 & 0.6838 & 0.5932 & 0.0000 \\
10 & LoRA  & RoBERTa-large-MNLI & 0.9940 & 0.0201 & 0.8897 & 0.5105 & 0.7396 \\
11 & Ours  & RoBERTa-large-MNLI & 0.9625 & 0.0983 & \textbf{0.8995} & 0.3400 & \textbf{0.7642} \\
12 & TLoRA & RoBERTa-large-MNLI & 0.7410 & 0.5449 & 0.7328 & 0.5784 & 0.3133 \\
\bottomrule
\end{tabular}
}
\caption{MRPC rank 32 results.}
\label{tab:mrpc_rank32}

\end{table}

\begin{table}[htbp]
\centering
\resizebox{\textwidth}{!}{
\begin{tabular}{cccccccc}
\toprule
ID & Method & Model & Tr Acc & Tr Loss & Val Acc & Val Loss & Val MCC \\
\midrule
1  & LoRA  & DeBERTa-base       & 0.9337 & 0.1948 & \textbf{0.8995} & 0.3200 & \textbf{0.7642} \\
2  & Ours  & DeBERTa-base       & 0.9188 & 0.2099 & 0.8824 & 0.3077 & 0.7314 \\
3  & TLoRA & DeBERTa-base       & 0.6767 & 0.5758 & 0.6838 & 0.5611 & 0.0000 \\
4  & LoRA  & OPT-125M           & 0.9905 & 0.0267 & 0.8015 & 0.9194 & 0.5253 \\
5  & Ours  & OPT-125M           & 0.9670 & 0.1004 & \textbf{0.8098} & 0.7045 & \textbf{0.5373} \\
6  & TLoRA & OPT-125M           & 0.7615 & 0.4990 & 0.7574 & 0.5119 & 0.3890 \\
7  & LoRA  & RoBERTa-base       & 0.9472 & 0.1454 & \textbf{0.8897} & 0.3320 & \textbf{0.7411} \\
8  & Ours  & RoBERTa-base       & 0.9226 & 0.1859 & 0.8848 & 0.3652 & 0.7283 \\
9  & TLoRA & RoBERTa-base       & 0.6832 & 0.5429 & 0.6838 & 0.5013 & 0.0000 \\
10 & LoRA  & RoBERTa-large-MNLI & 0.9938 & 0.0201 & 0.8873 & 0.5209 & 0.7335 \\
11 & Ours  & RoBERTa-large-MNLI & 0.9716 & 0.0739 & \textbf{0.8897} & 0.4930 & \textbf{0.7400} \\
12 & TLoRA & RoBERTa-large-MNLI & 0.8390 & 0.3601 & 0.8578 & 0.3847 & 0.6604 \\
\bottomrule
\end{tabular}
}
\caption{MRPC rank 64 results.}
\label{tab:mrpc_rank64}

\end{table}

\newpage
\section{The Validation Accuracy Trends of Different Methods}
\label{appendix:val acc qnli}
The figures below illustrate the validation accuracy trends on the QNLI dataset, comparing various models tuned with our method, LoRA and TLoRA across ranks of 8, 16, 32 and 64.

\begin{figure}[htbp]
	\centering
	\begin{subfigure}{0.48\linewidth}
		\centering
		\includegraphics[width=\linewidth]{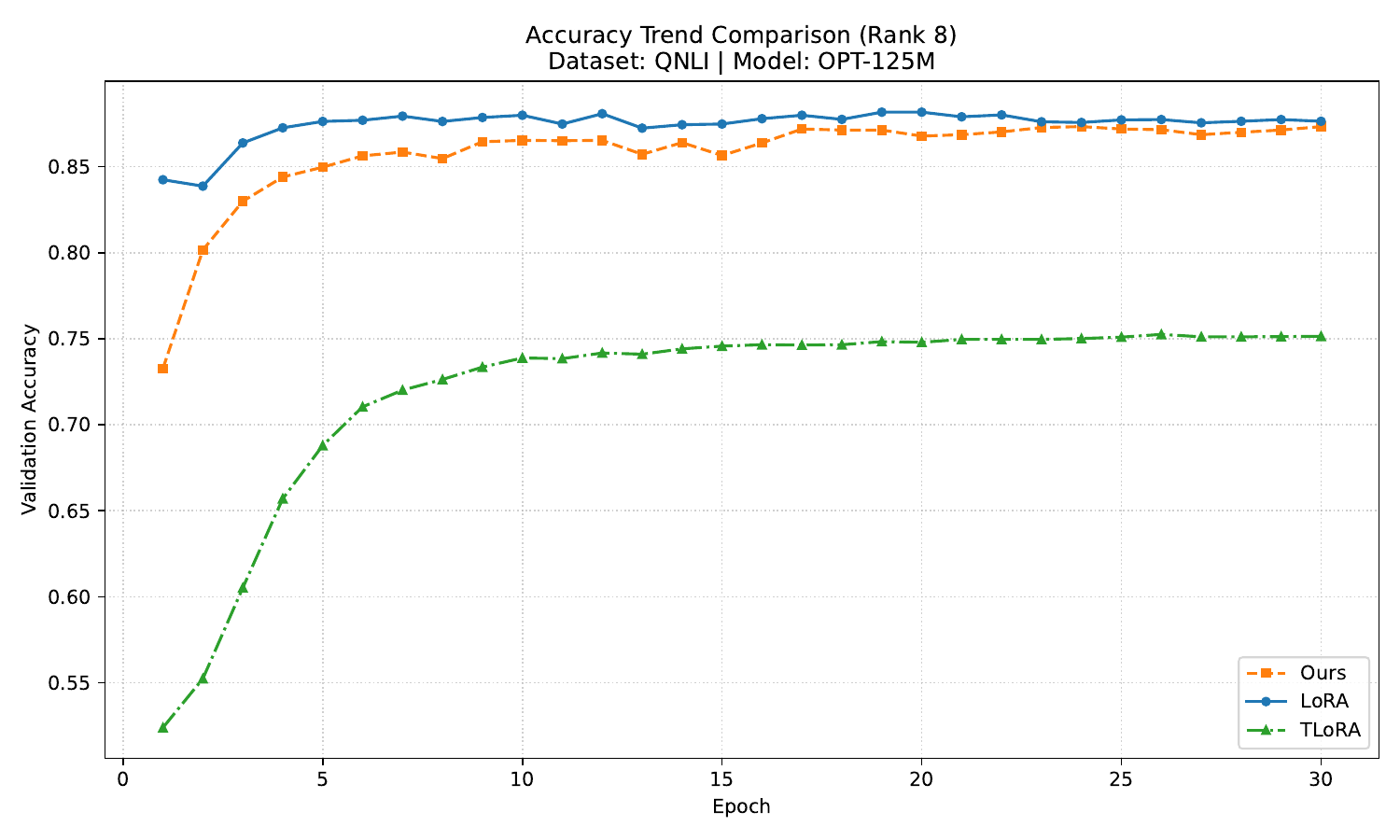}
	\end{subfigure}
	\hfill
	\begin{subfigure}{0.48\linewidth}
		\centering
		\includegraphics[width=\linewidth]{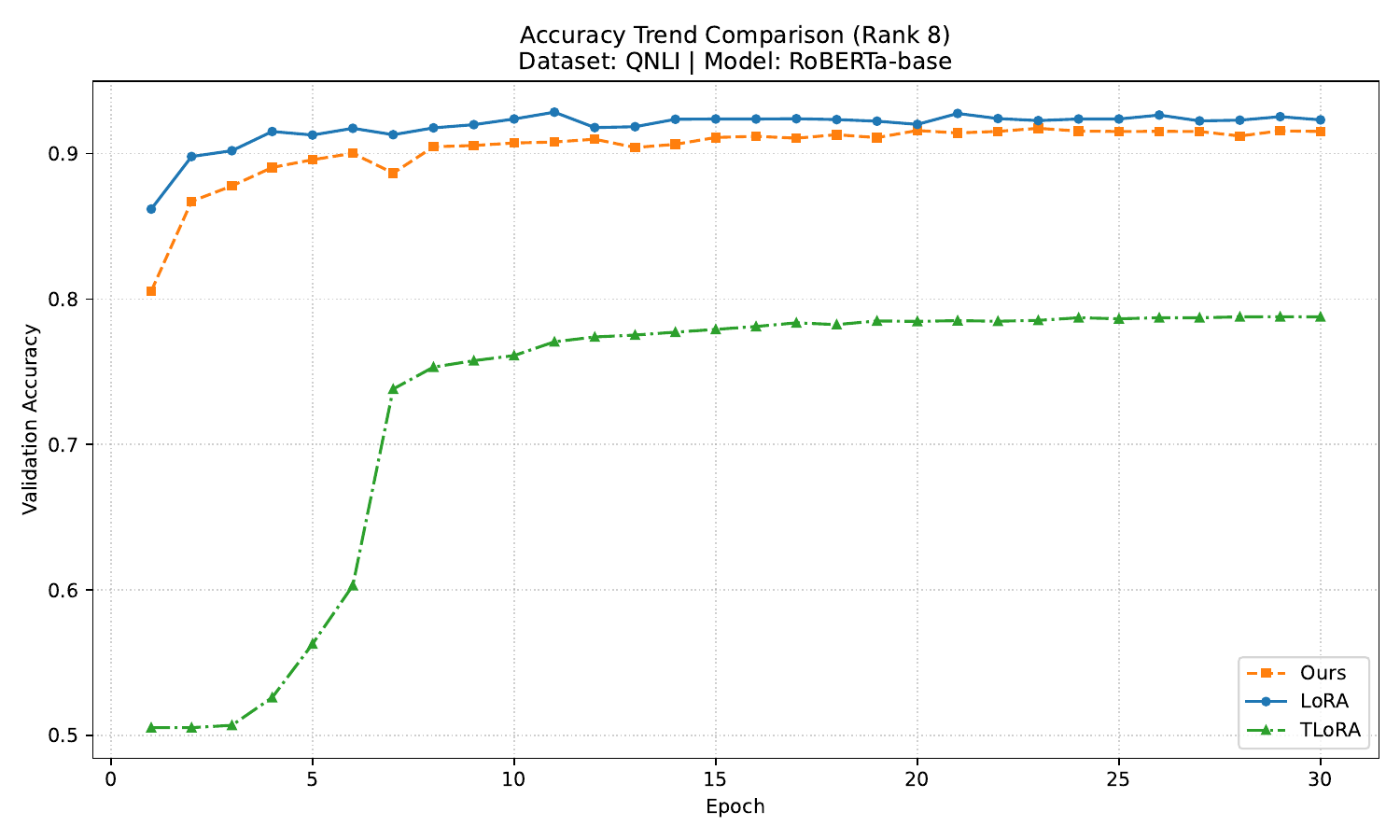}
	\end{subfigure}
	
	\vspace{0.5em}
	
	\begin{subfigure}{0.48\linewidth}
		\centering
		\includegraphics[width=\linewidth]{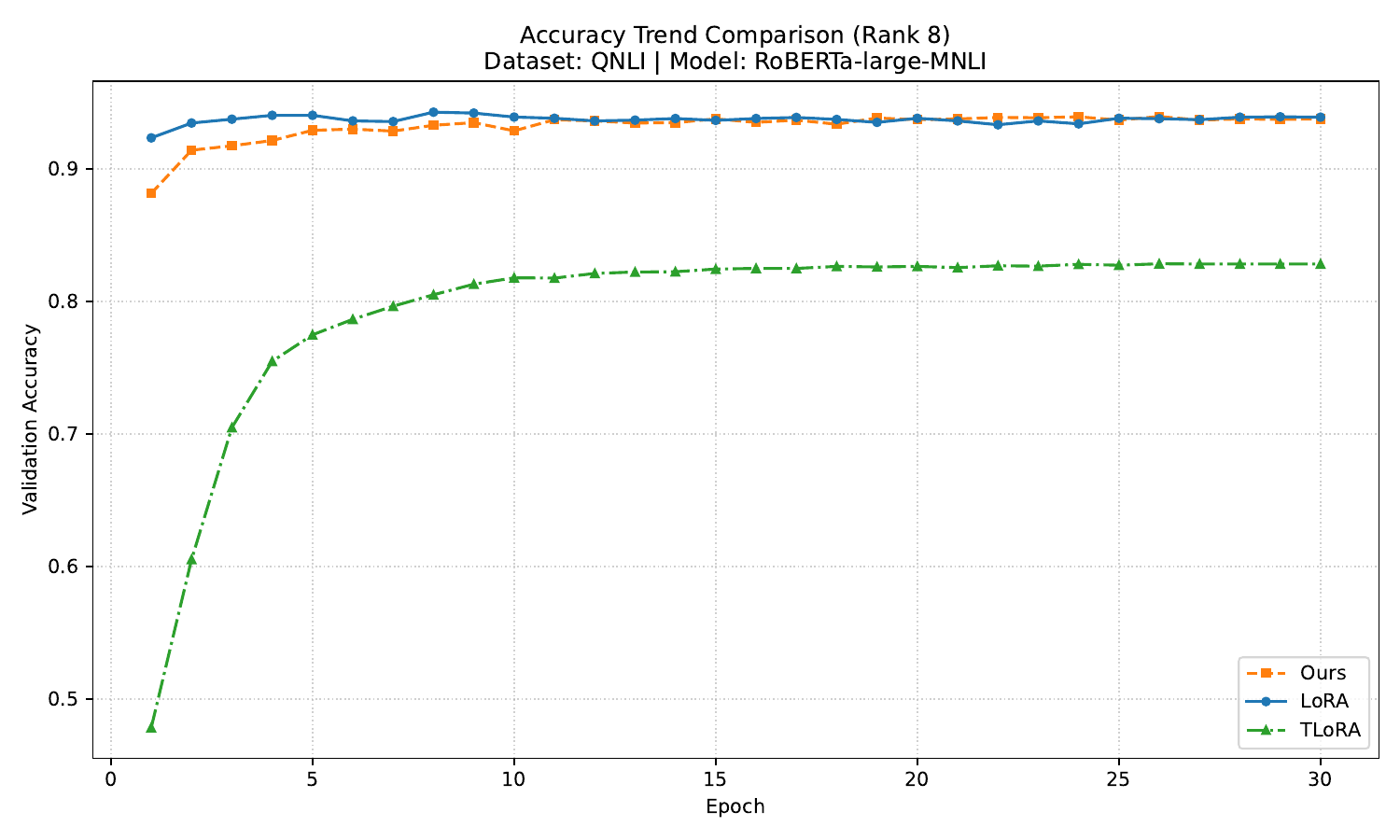}
	\end{subfigure}
	\hfill
	\begin{subfigure}{0.48\linewidth}
		\centering
		\includegraphics[width=\linewidth]{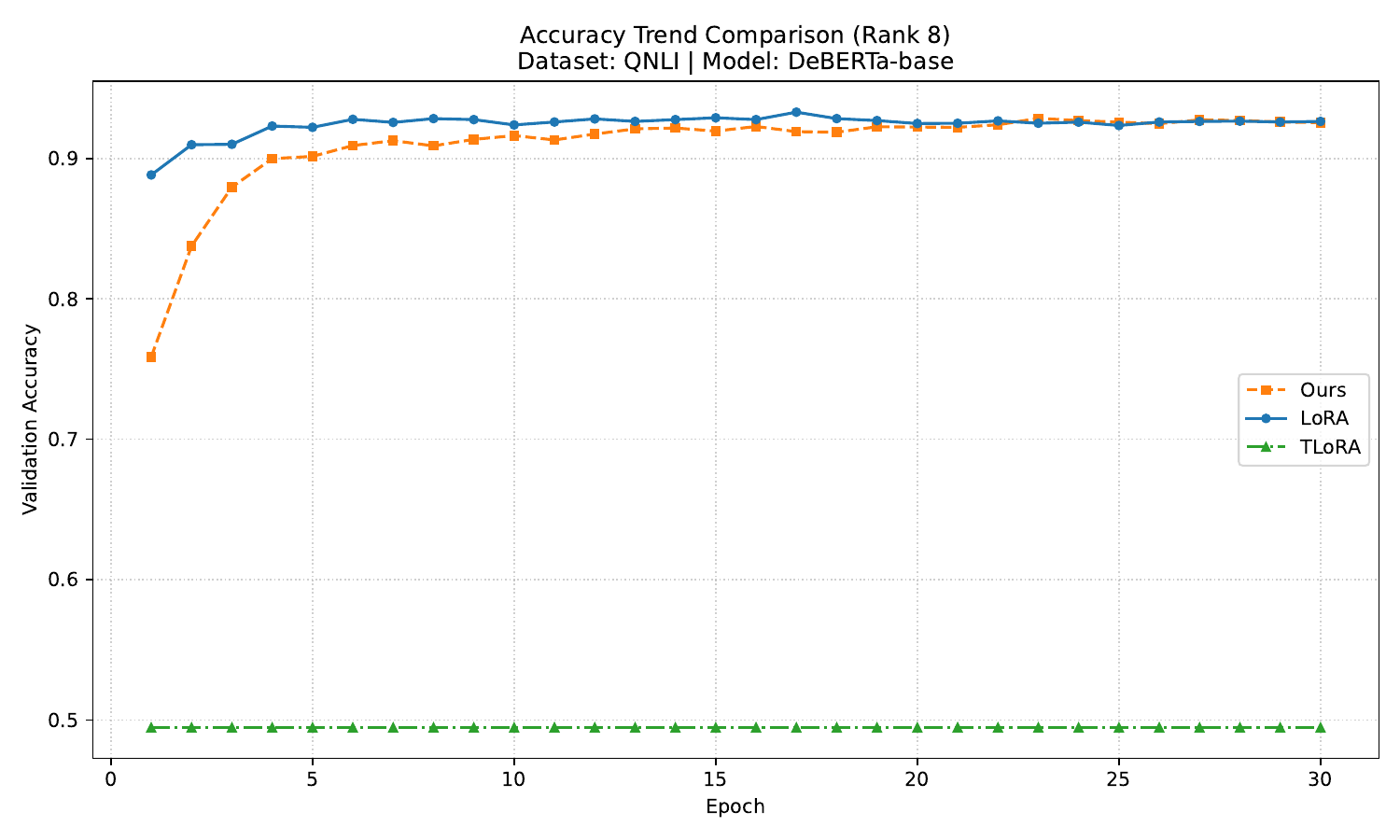}
	\end{subfigure}
	
	\caption{
    		Validation accuracy on QNLI at rank 8 across different backbone models.
		Each subplot compares three methods.
	}
\end{figure}

\begin{figure}[htbp]
	\centering
	\begin{subfigure}{0.48\linewidth}
		\centering
		\includegraphics[width=\linewidth]{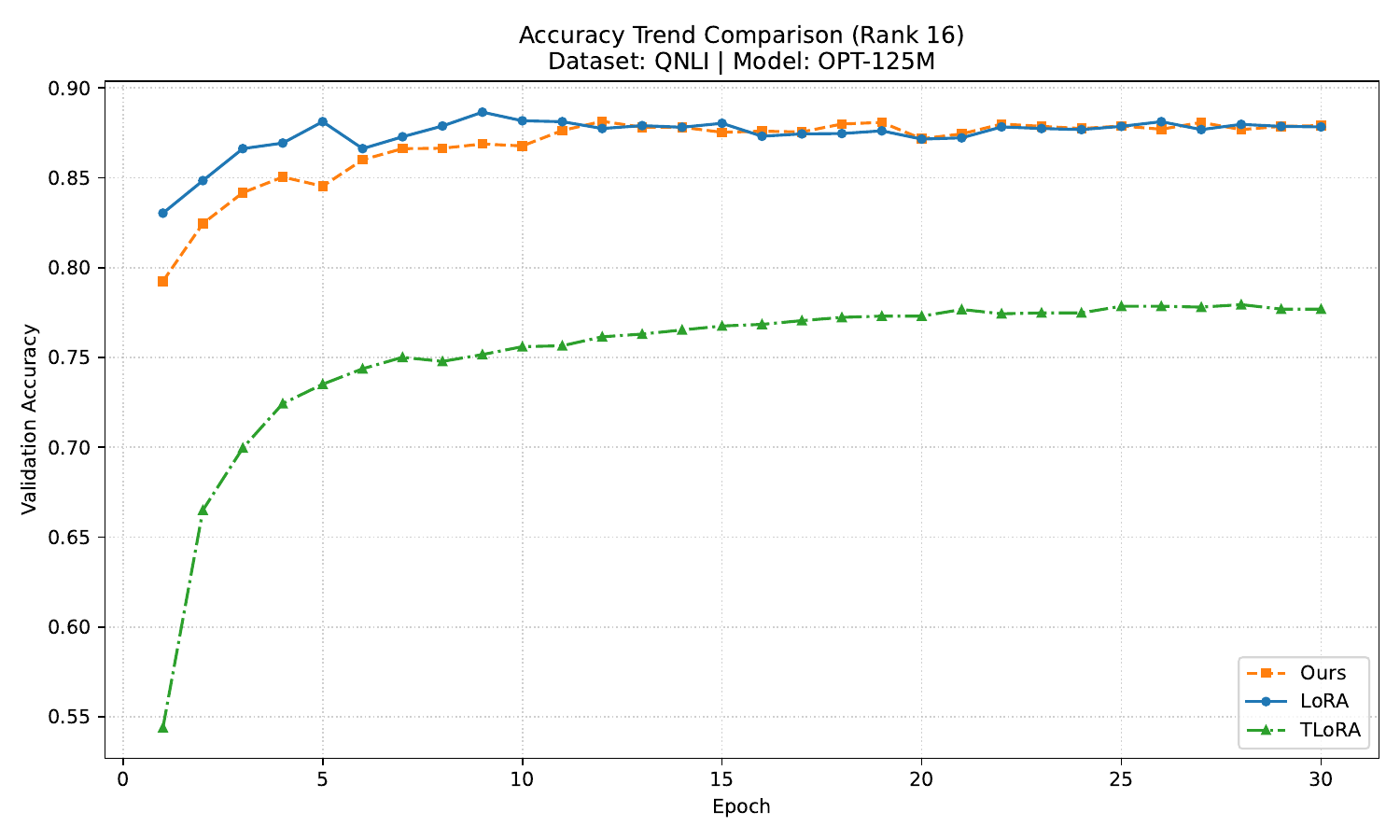}
	\end{subfigure}
	\hfill
	\begin{subfigure}{0.48\linewidth}
		\centering
		\includegraphics[width=\linewidth]{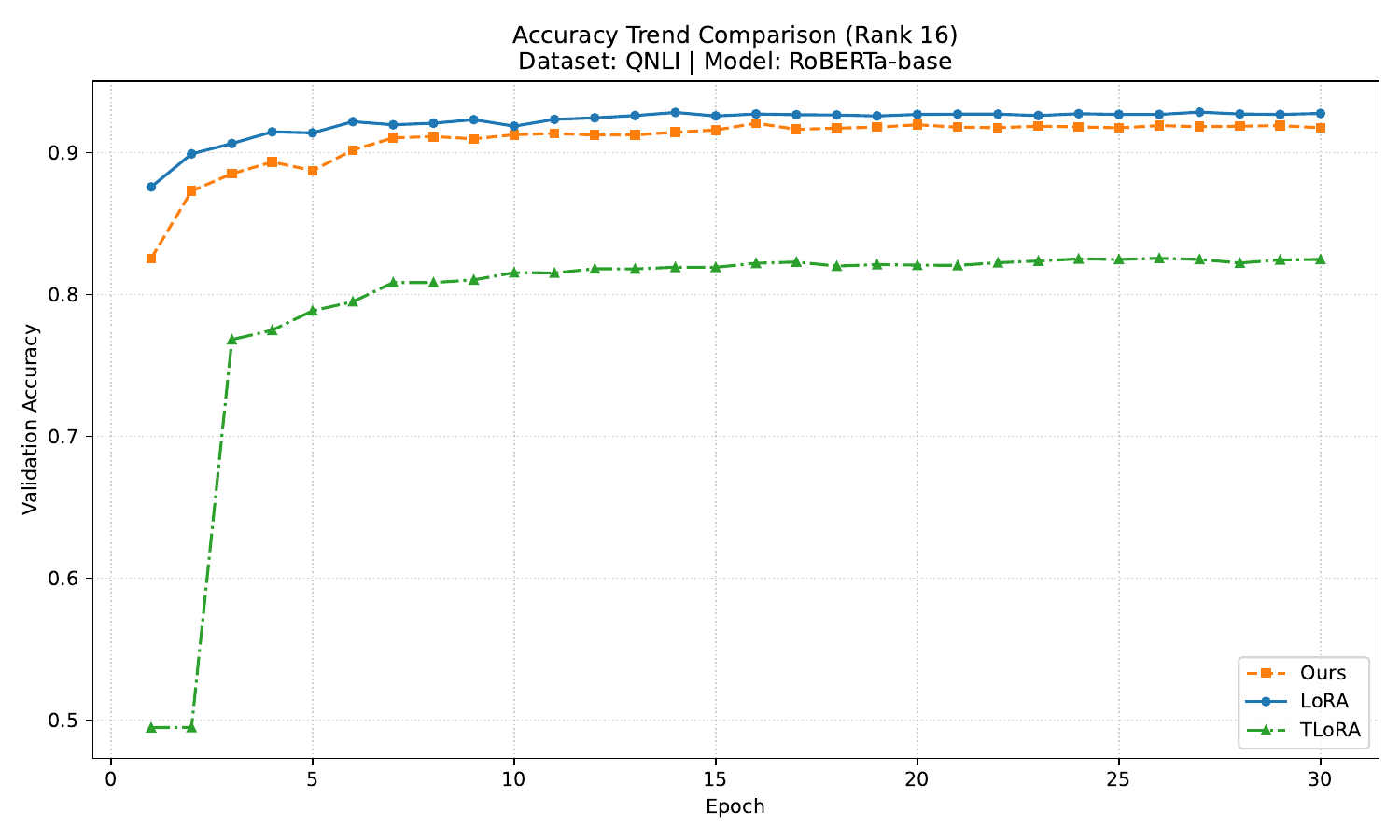}
	\end{subfigure}
	
	\vspace{0.5em}
	
	\begin{subfigure}{0.48\linewidth}
		\centering
		\includegraphics[width=\linewidth]{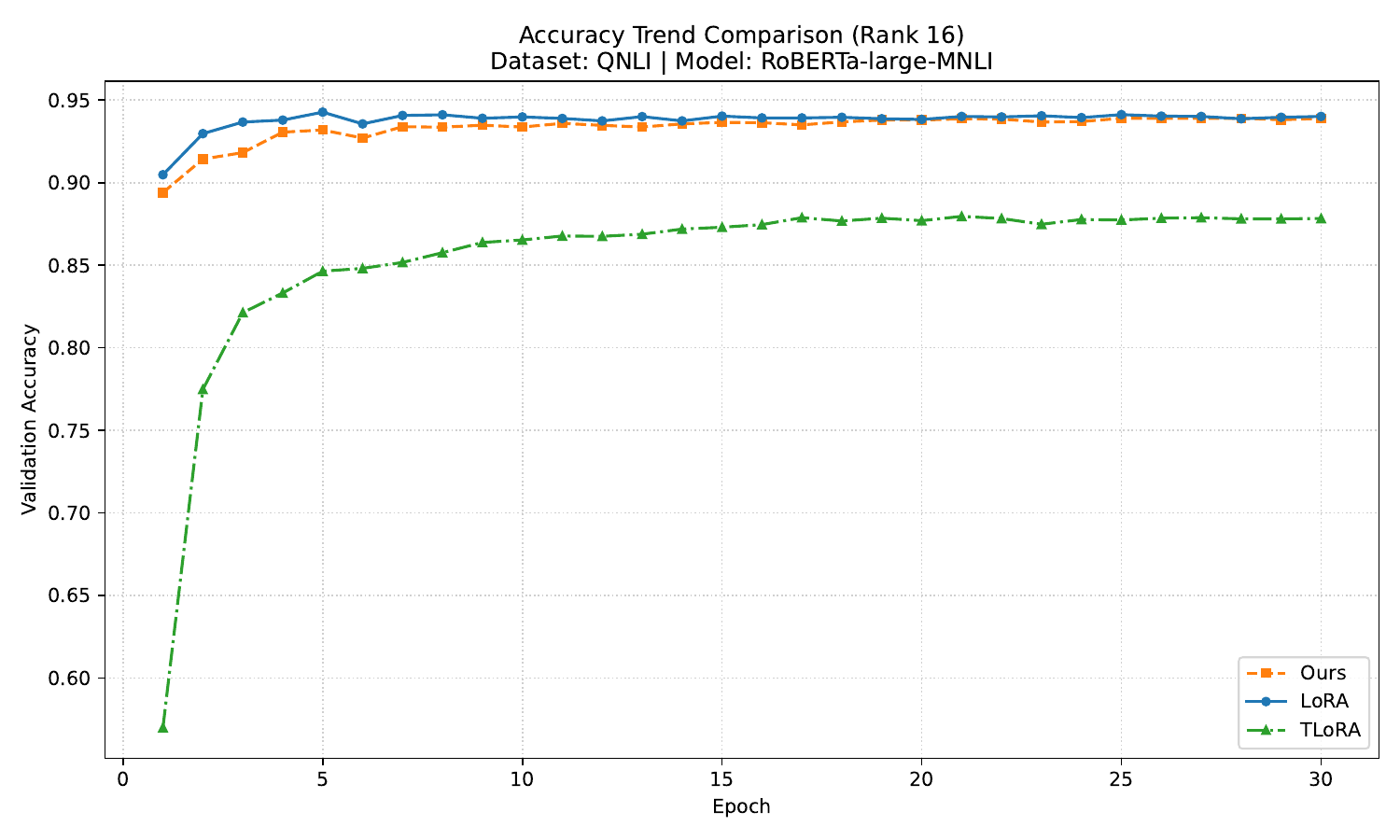}
	\end{subfigure}
	\hfill
	\begin{subfigure}{0.48\linewidth}
		\centering
		\includegraphics[width=\linewidth]{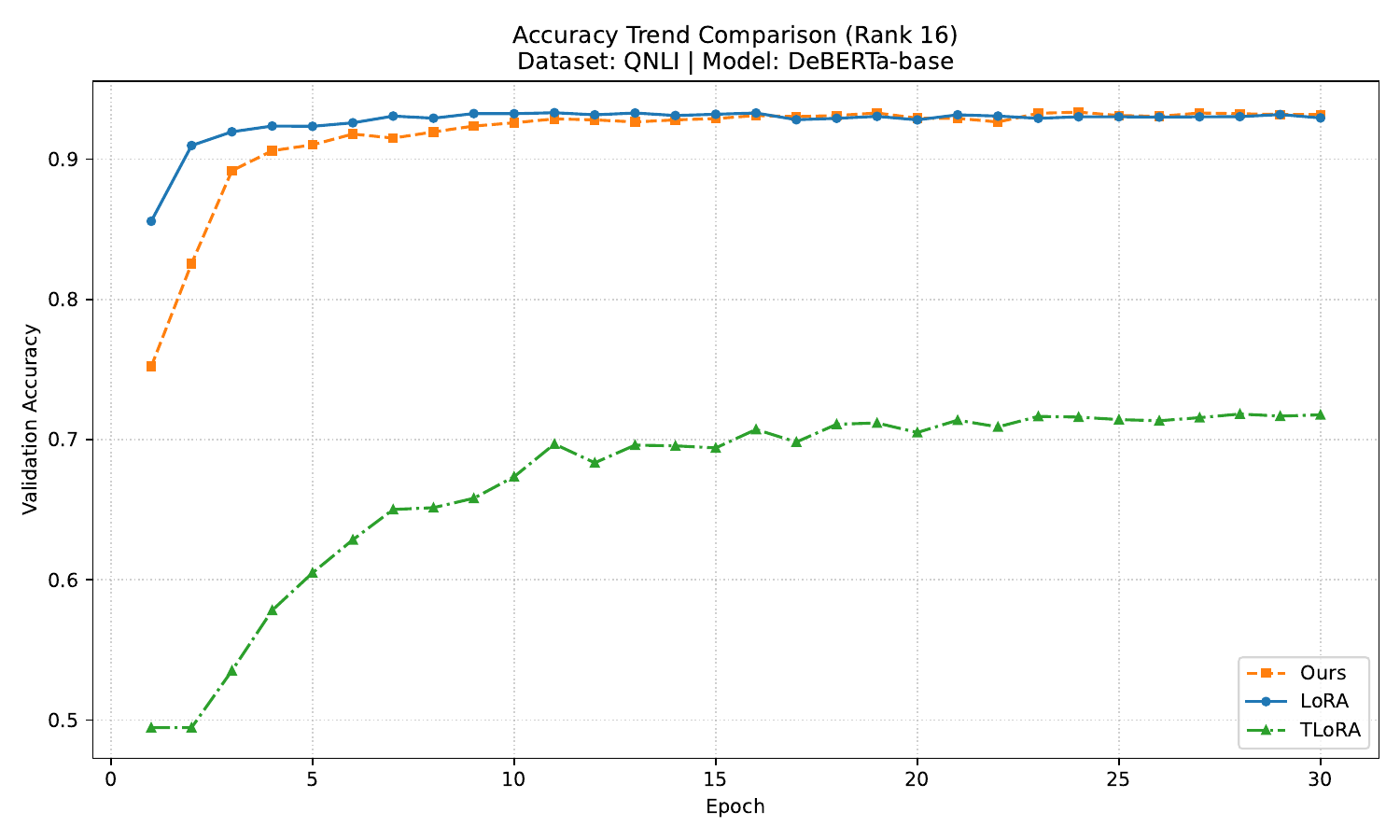}
	\end{subfigure}
	
	\caption{
		Validation accuracy on QNLI at rank 16 across different backbone models.
		Each subplot compares three methods.
	}
\end{figure}

\begin{figure}[htbp]
	\centering
	\begin{subfigure}{0.48\linewidth}
		\centering
		\includegraphics[width=\linewidth]{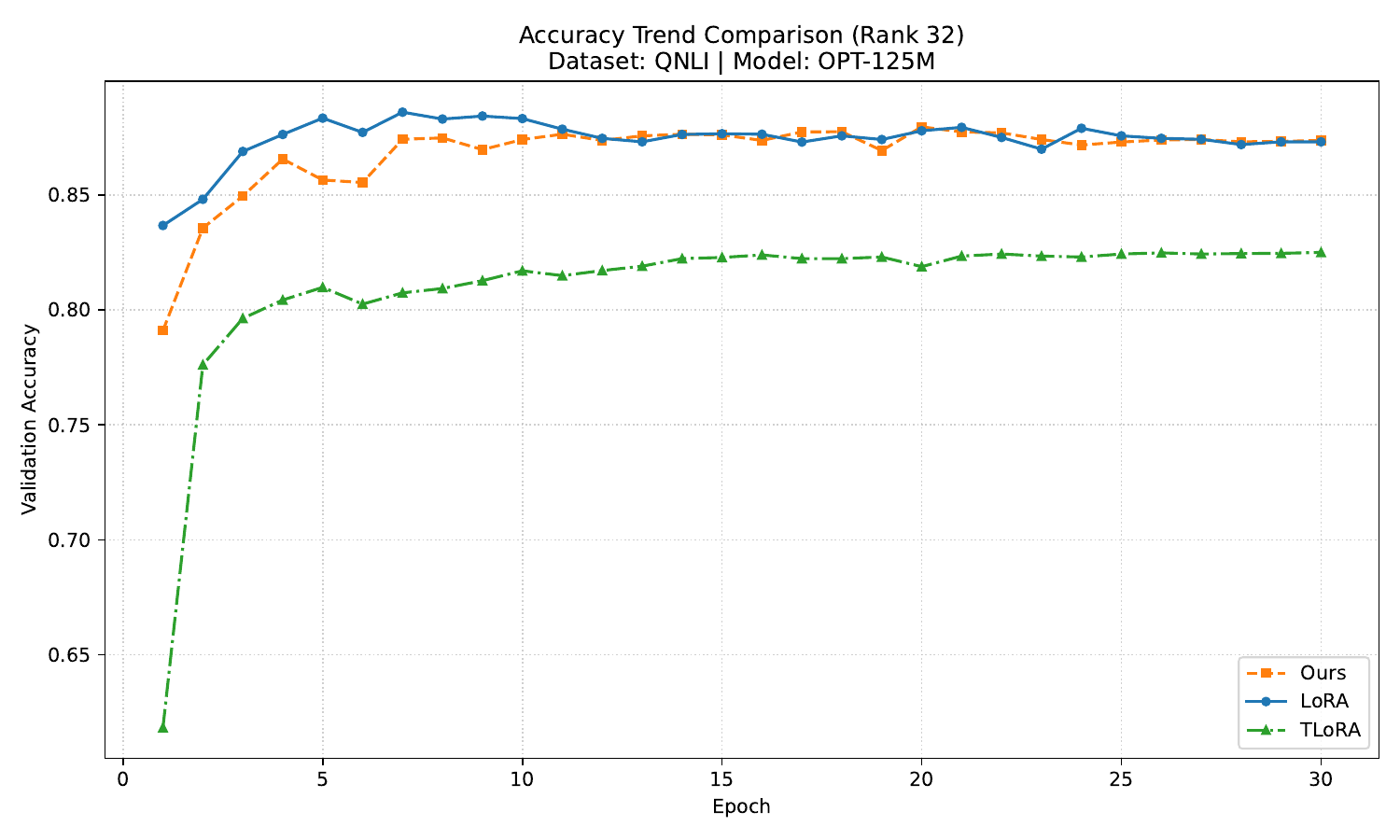}
	\end{subfigure}
	\hfill
	\begin{subfigure}{0.48\linewidth}
		\centering
		\includegraphics[width=\linewidth]{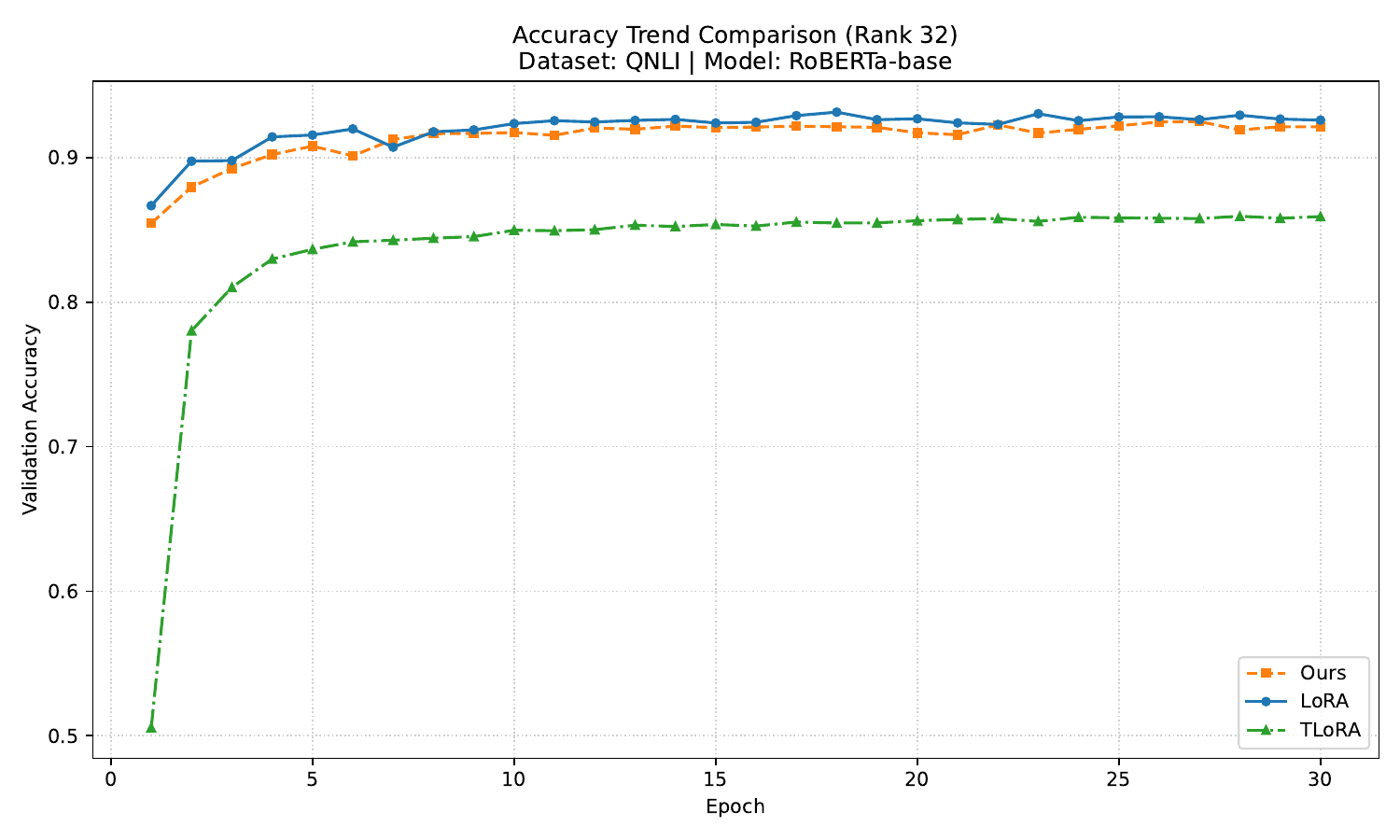}
	\end{subfigure}
	
	\vspace{0.5em}
	
	\begin{subfigure}{0.48\linewidth}
		\centering
		\includegraphics[width=\linewidth]{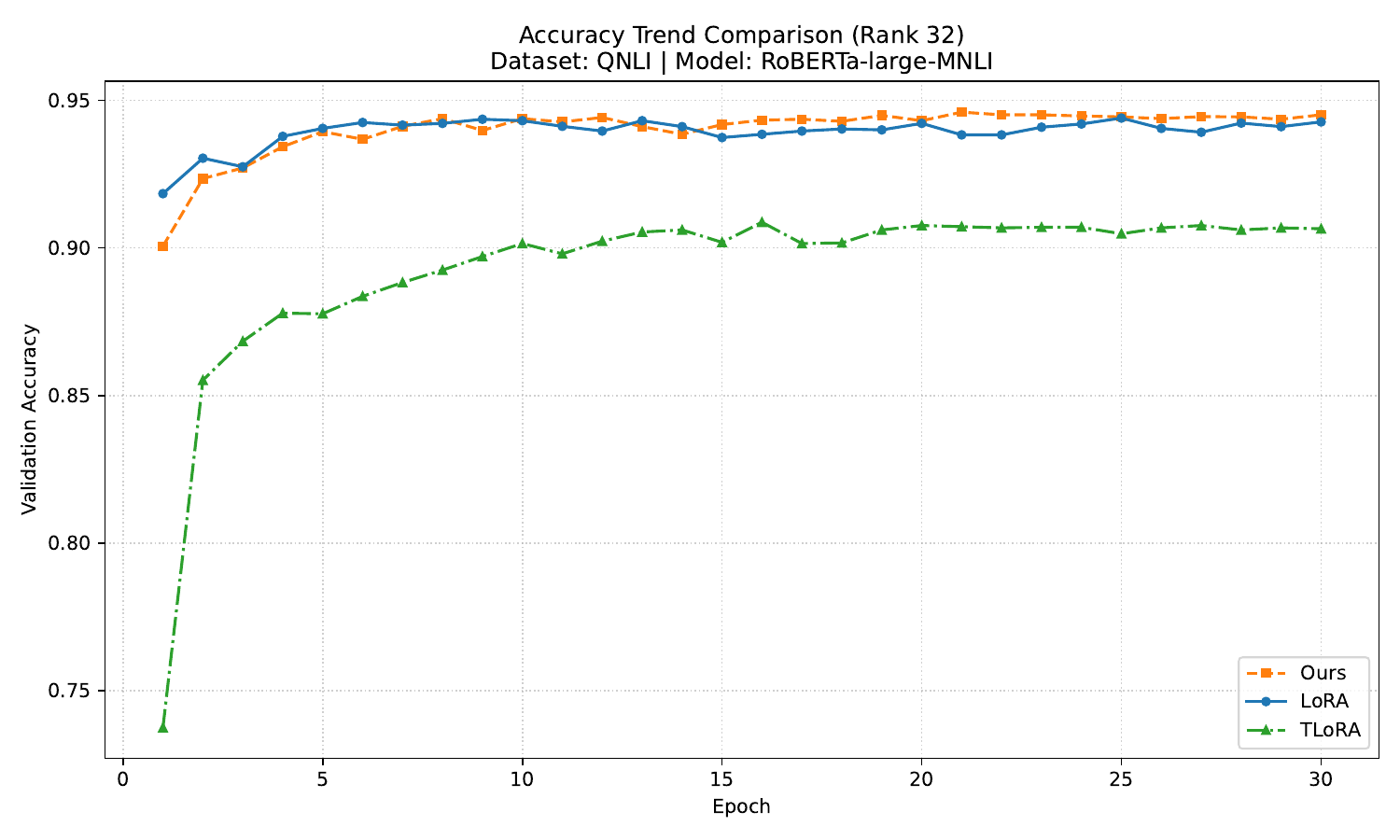}
	\end{subfigure}
	\hfill
	\begin{subfigure}{0.48\linewidth}
		\centering
		\includegraphics[width=\linewidth]{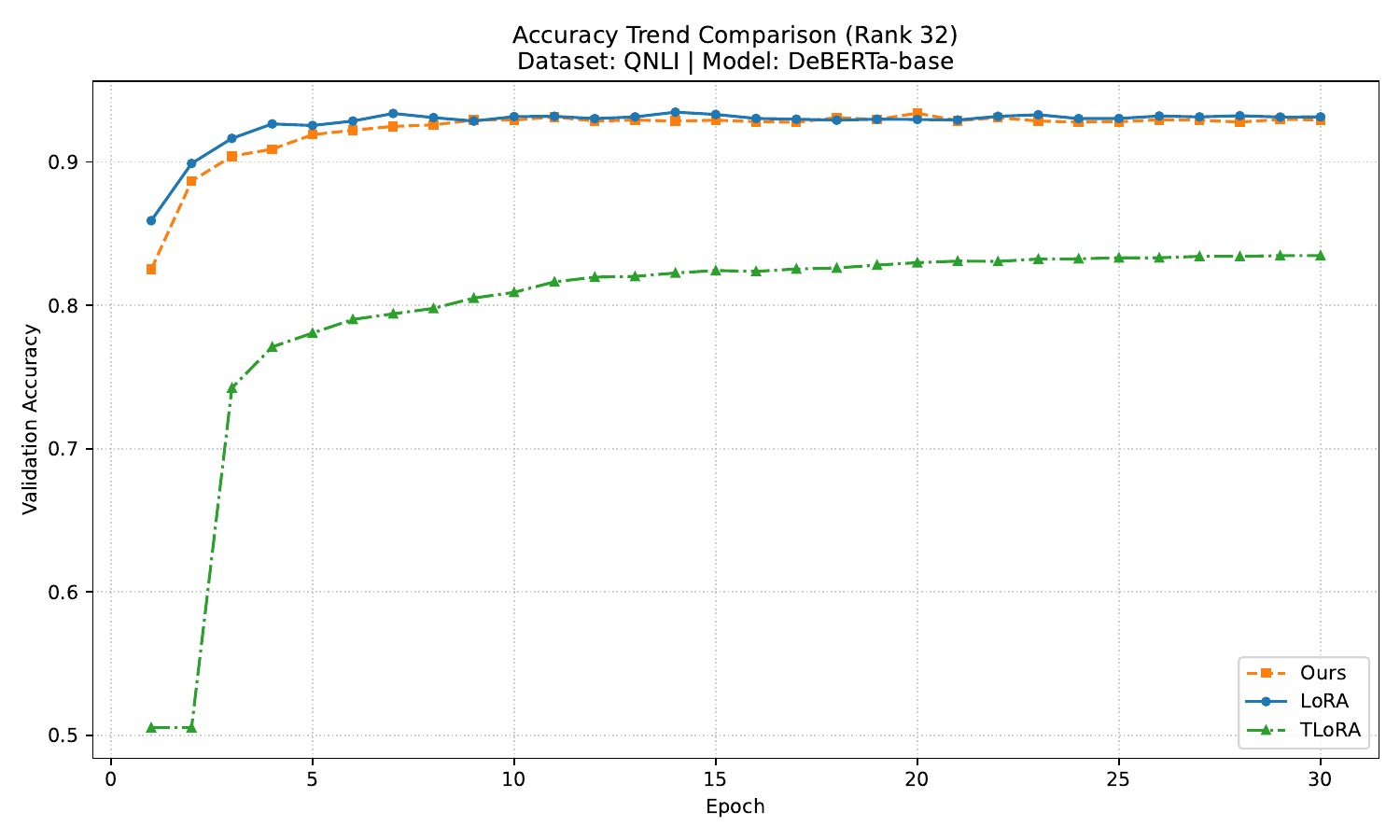}
	\end{subfigure}
	
	\caption{
		Validation accuracy on QNLI at rank 32 across different backbone models.
		Each subplot compares three methods.
	}
\end{figure}

\begin{figure}[htbp]
	\centering
	\begin{subfigure}{0.48\linewidth}
		\centering
		\includegraphics[width=\linewidth]{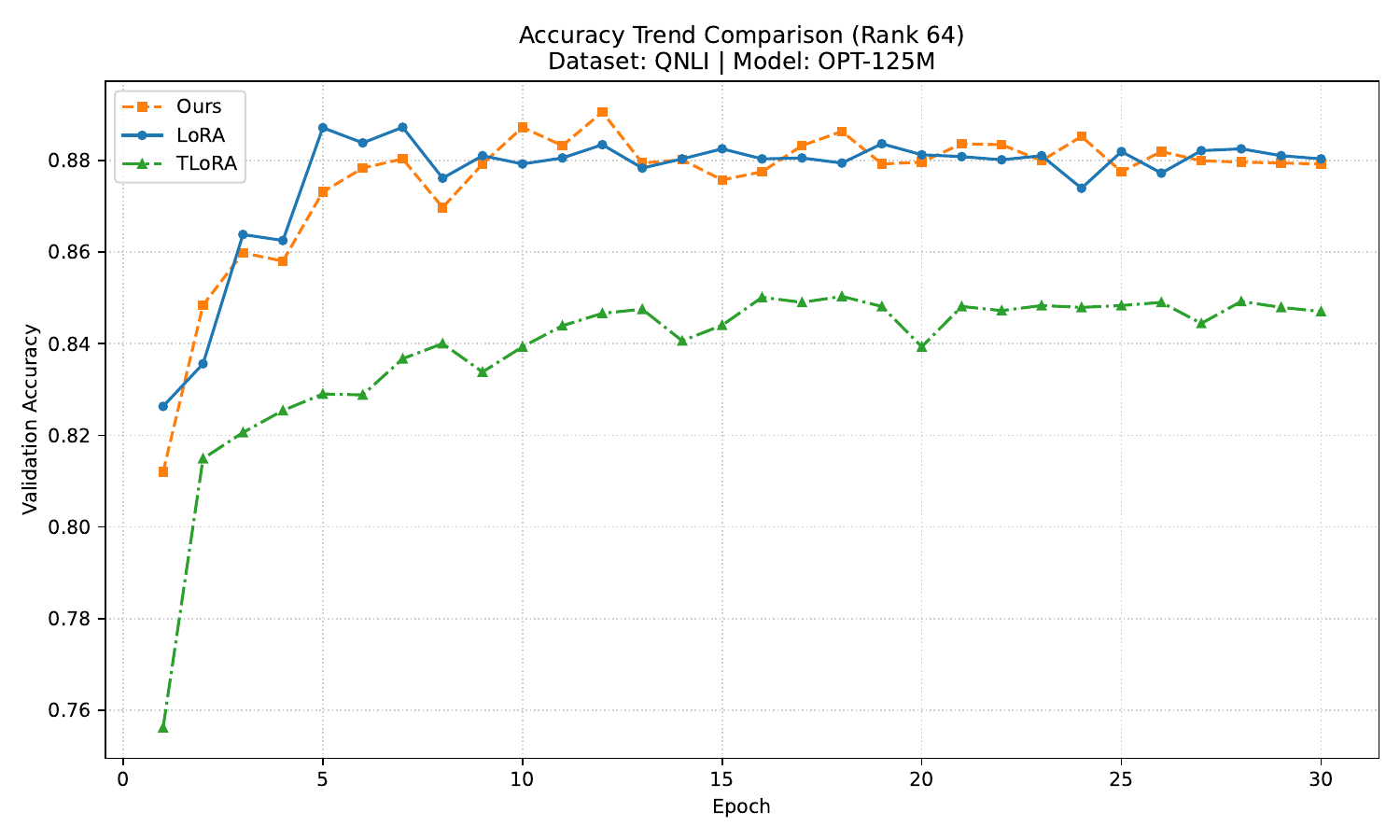}
	\end{subfigure}
	\hfill
	\begin{subfigure}{0.48\linewidth}
		\centering
		\includegraphics[width=\linewidth]{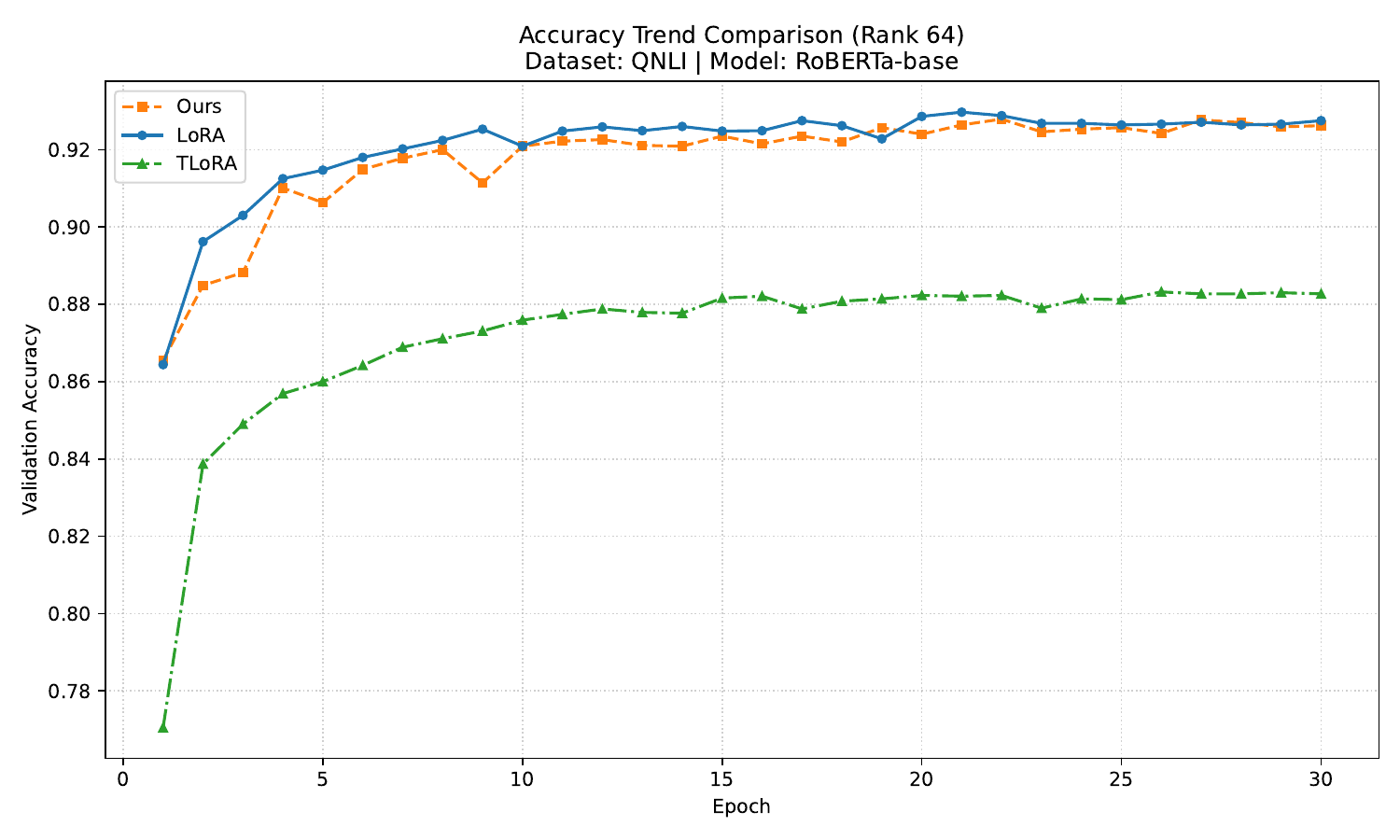}
	\end{subfigure}
	
	\vspace{0.5em}
	
	\begin{subfigure}{0.48\linewidth}
		\centering
		\includegraphics[width=\linewidth]{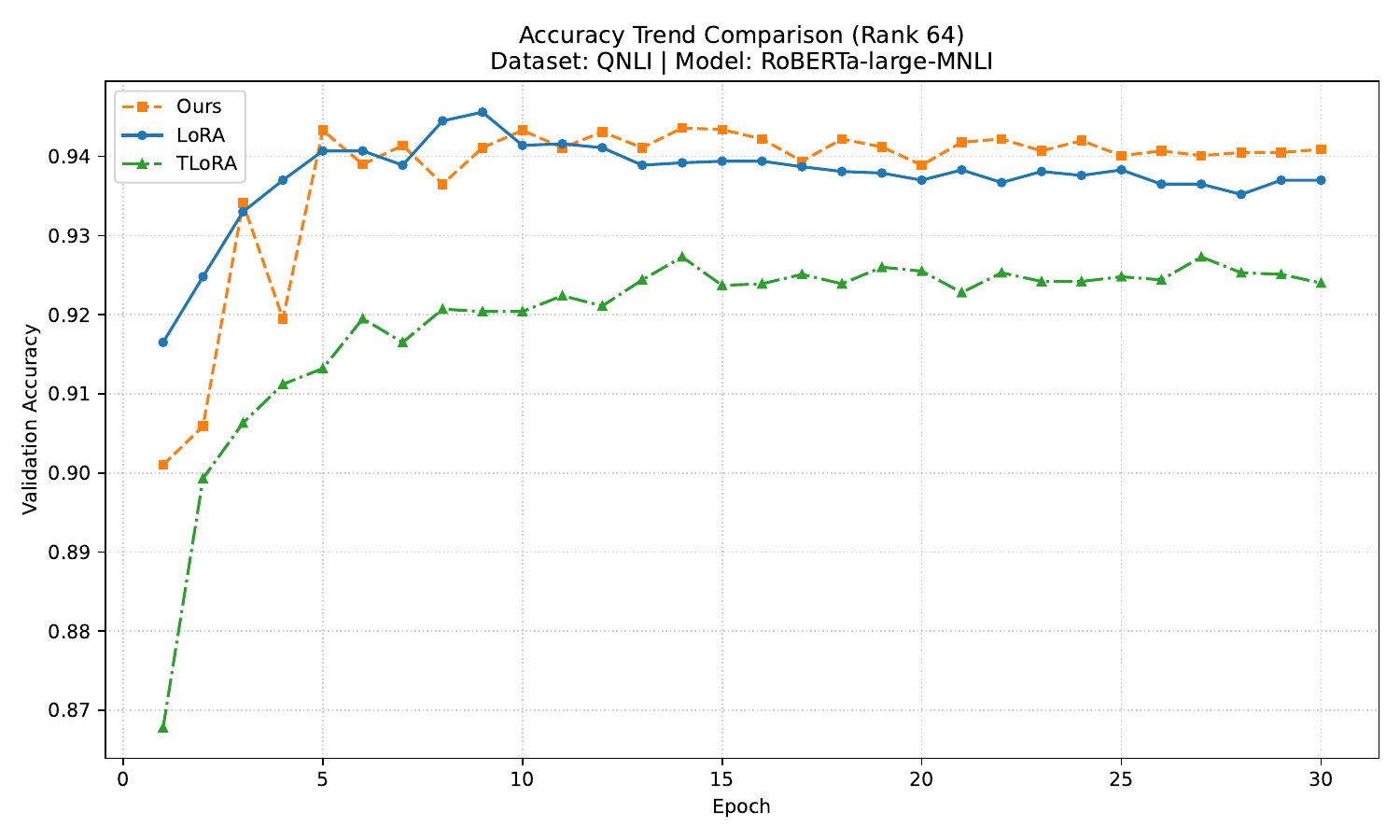}
	\end{subfigure}
	\hfill
	\begin{subfigure}{0.48\linewidth}
		\centering
		\includegraphics[width=\linewidth]{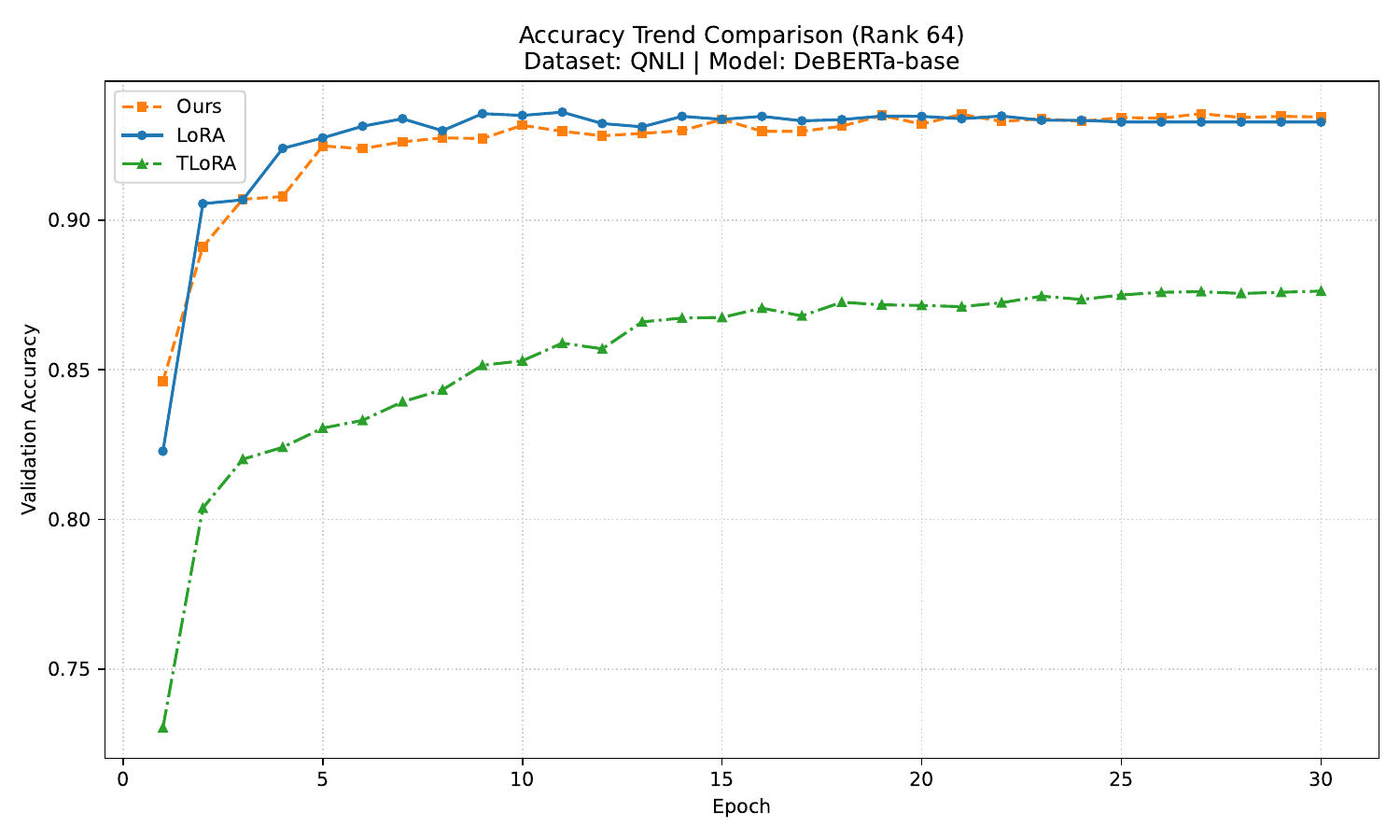}
	\end{subfigure}
	
	\caption{
		Validation accuracy on QNLI at rank 64 across different backbone models.
		Each subplot compares three methods.
	}
\end{figure}

\newpage
\section{More Results for Ratio Choice}
\label{appendix:ratio}
\begin{figure}[htbp]
	\centering
	\begin{subfigure}{0.48\linewidth}
		\centering
		\includegraphics[width=\linewidth]{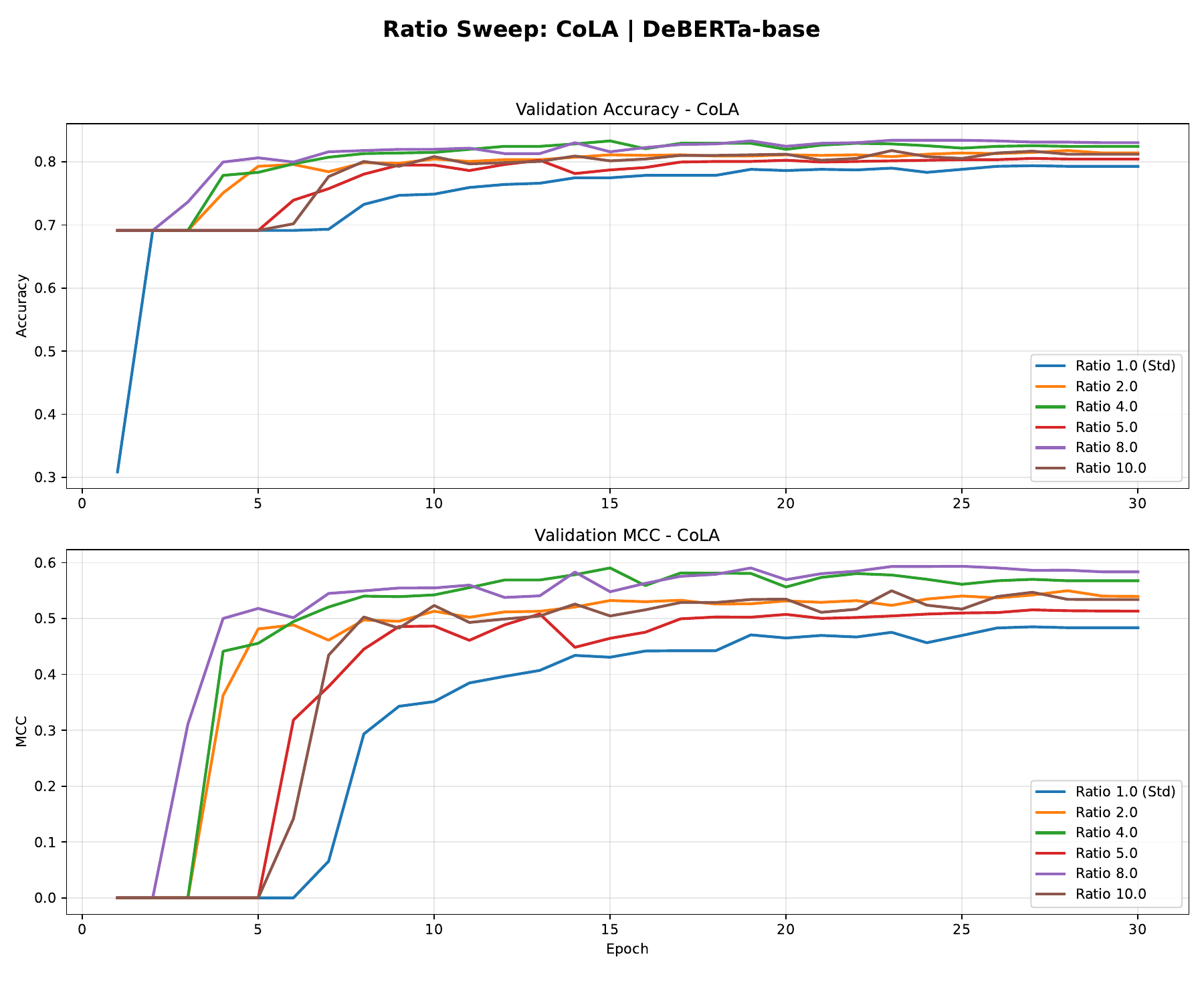}
	\end{subfigure}
	\hfill
	\begin{subfigure}{0.48\linewidth}
		\centering
		\includegraphics[width=\linewidth]{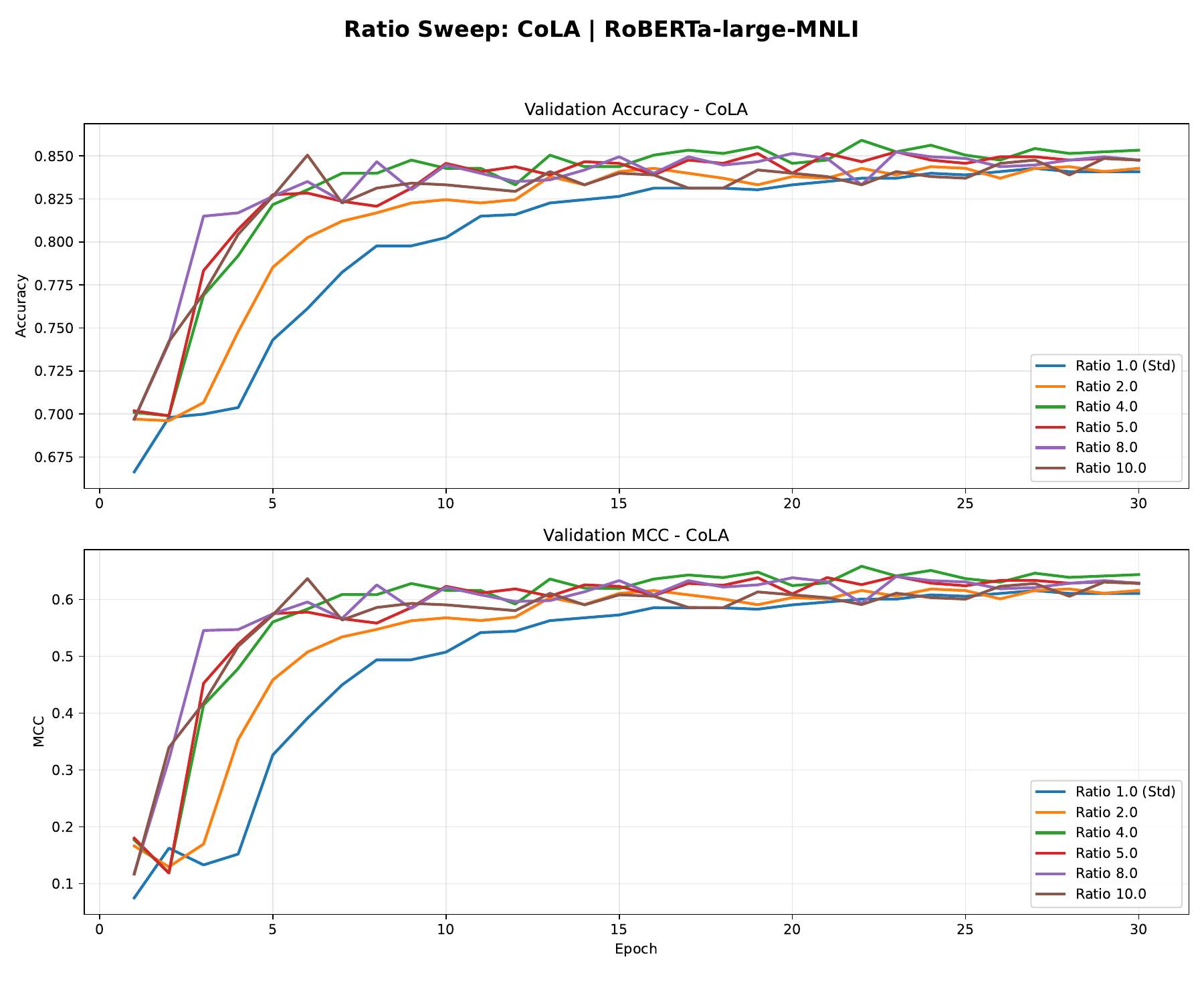}
	\end{subfigure}	
	\caption{
    		Validation accuracy and MCC on CoLA at rank 8 across different models.
	}
    \label{Fig:ratio_cola}
\end{figure}

Figure~\ref{Fig:ratio_cola} illustrates the validation accuracy and MCC trends on the CoLA dataset, comparing various models tuned with our method across ratio choices of 1.0, 2.0, 4.0, 5.0, 8.0 and 10.0.

\end{document}